\documentclass{article}

\PassOptionsToPackage{numbers}{natbib}
\usepackage[preprint]{neurips_2020}
\usepackage[utf8]{inputenc}
\usepackage{amsmath}
\usepackage{amsfonts}
\usepackage{amssymb}
\usepackage{amsthm}
\usepackage{mathrsfs}
\usepackage{graphbox}
\usepackage{graphicx}

\DeclareMathOperator*{\argmin}{arg\,min}

\usepackage{algorithmic}
\newcommand{\algcomment}[1]{\(\triangleright\) #1}

\usepackage[ruled]{algorithm}

\usepackage{authblk}

\newtheorem{definition}{Definition}

\usepackage{booktabs}
\usepackage{multirow}
\usepackage{threeparttable}
\usepackage{siunitx}
\usepackage[title]{appendix}
\usepackage{chngcntr}

\title{Analysis of Dominant Classes in Universal Adversarial Perturbations}

\author[1]{Jon Vadillo}
\author[1]{Roberto Santana}
\author[1,2]{Jose A. Lozano}

\affil[1]{Department of Computer Science and Artificial Intelligence, \protect\\ University of the Basque Country UPV/EHU. \authorcr
  \{\tt jon.vadillo, roberto.santana, ja.lozano\}@ehu.eus}
\affil[2]{Basque Center for Applied Mathematics (BCAM). \authorcr
  \tt jlozano@bcamath.org}

\begin{document}

\maketitle

\begin{abstract}
The reasons why Deep Neural Networks are susceptible to being fooled by adversarial examples remains an open discussion. Indeed, many different strategies can be employed to efficiently generate adversarial attacks, some of them relying on different theoretical justifications. Among these strategies, universal (input-agnostic) perturbations are of particular interest, due to their capability to fool a network independently of the input in which the perturbation is applied. In this work, we investigate an intriguing phenomenon of universal perturbations, which has been reported previously in the literature, yet without a proven justification: universal perturbations change the predicted classes for most inputs into one particular (dominant) class, even if this behavior is not specified during the creation of the perturbation. In order to justify the cause of this phenomenon, we propose a number of hypotheses and experimentally test them using a speech command classification problem in the audio domain as a testbed. Our analyses reveal interesting properties of universal perturbations, suggest new methods to generate such attacks and provide an explanation of dominant classes, under both a geometric and a data-feature perspective.
\end{abstract}

\section{Introduction}

Universal adversarial perturbations  \cite{moosavi-dezfooli2017universal} are input-agnostic perturbations capable of fooling a Deep Neural Network (DNN) while remaining imperceptible for humans. These perturbations are generally created as \textit{untargeted} attacks, so that no preference over the (incorrect) output class is assumed \cite{moosavi-dezfooli2017universal,mopuri2017fast,khrulkov2018art,vadillo2019universal}. However, previous work \cite{moosavi-dezfooli2017universal, co2020universal, hirano2020universal, behjati2019universal} has reported a phenomenon regarding the effect of universal perturbations in the attacked model: the preference of the perturbation to change the class of the inputs into a particular \textit{dominant} class, without this being specified or imposed in the generation of the perturbation. Thus, some classes (or class regions in the decision space) act as \textit{attractors} under the effect of universal perturbations.

In this paper, we analyze this phenomenon with the aim of sheding light on the (still misunderstood) vulnerability of DNNs to universal perturbations. The main contributions of our paper are the following:
\begin{itemize}
    \item First, we propose a number of hypotheses to explain and characterize the existence of dominant classes linked to universal adversarial perturbations, and revisit previous hypotheses and open questions in the related work.
    \item We experimentally test the proposed hypotheses using a speech command classification task in the audio domain as a testbed. To the best of our knowledge, this is the first work in which the analysis of dominant classes is studied for the audio domain. Apart from providing evidence of the validity of the proposed hypotheses, our results reveal interesting properties of the DNN sensitivity to different types of perturbations. 
    \item Finally, we highlight a number of differences between the image domain and the audio domain regarding the analysis of adversarial examples, contributing to a more general understanding of adversarial machine learning.
\end{itemize}

\section{Related work}
\label{sec:related_work}
Universal adversarial perturbations for DNNs were introduced in \cite{moosavi-dezfooli2017universal} for image classification tasks. The goal of such perturbations is to fool a DNN for ``most'' natural inputs when they are applied to them, and, at the same time, to be imperceptible for humans. Formally, following the notation used in \cite{moosavi-dezfooli2017analysis}, a perturbation $v$ is said to be $(\xi,\delta)$-universal if the following conditions are satisfied:
\begin{align}
    & ||v||_2 \leq \xi, \\
    &\mathbb{P}_{x\sim\mu}\left[ f(x+v)\neq f(x) \right] \geq 1 - \delta, 
\end{align}
being $\mu$ the distribution of natural inputs in the $d$-dimensional input space $\mathbb{R}^d$, and $f(x)$ the output class assigned to an input $x$ by a classifier  ${f:\mathbb{R}^d\rightarrow \{y_1,...,y_k\}}$. 

The discovery of such attacks for state-of-the-art DNNs has led to a deeper study of their properties. In \cite{moosavi-dezfooli2017universal}, the vulnerability of DNNs to universal perturbations is empirically studied in the image domain, which is attributed in part to the geometry of the decision boundaries learned by the DNNs. In particular, it is shown that, in the vicinity of natural inputs, perturbations normal to the decision boundaries are \textit{correlated}, in the sense that they approximately span a low dimensional subspace (in comparison to the dimensionality of the input space). Thus, being 
\begin{equation}
\label{eq:optimal_adv}
v_x = \argmin_v ||v||_2 \ \text{ s.t. } \ f(x) \neq f(x+v)
\end{equation}
the minimal perturbation capable of changing the output of an input $x$ (hence \textit{normal} to the decision boundary at $x+v_x$), it is possible to find a subspace $S \subset X$, with $dim(S)\ll dim(X)$, so that $v_x\in S$ for $x \sim \mu$.
The existence of such a subspace implies that even random perturbations (with small norms) sampled from $S$ are likely to cause a misclassification for a large number of inputs \cite{moosavi-dezfooli2017universal}. 
This hypothesis is further developed in \cite{moosavi-dezfooli2017analysis}, also for the image domain, where the vulnerability of classifiers to universal perturbations is formalized, under the assumption of locally linear decision boundaries in the vicinity of natural inputs. An illustration of a linear approximation of the decision boundary is shown in Figure \ref{fig:boundary_models} (left).

However, the assumption of locally linear decision boundaries becomes insufficient to comprehensively formalize the vulnerability of DNNs to universal perturbations. Indeed, there is a crucial connection between that vulnerability and the curvature of the decision boundaries \cite{moosavi-dezfooli2017analysis}: there exist common perturbation directions (i.e., span a low-dimensional subspace) in the input space for which, starting from natural inputs, the decision boundaries are positively curved along these directions. See Figure \ref{fig:boundary_models} (right) for a comparison between a positively curved boundary and a negatively curved boundary. The positive curvature of the decision boundaries implies small upper bounds for the amount of perturbation required to surpass the decision boundaries, as depicted in Figure \ref{fig:boundary_models} (right). Thus, those positive curvatures increase the vulnerability of DNNs, as smaller perturbations are required to fool the model. At the same time, the fact that those directions are also \textit{common} for multiple inputs implies the existence of small \emph{input-agnostic} adversarial perturbations.

In a further analysis developed in \cite{jetley2018friends}, it is shown that the directions in the input space for which the decision boundaries are highly curved are indeed associated by the DNN with class identities (the further we move in one of such directions, the higher - or lower- the confidence of the model in one particular class is). Moreover, it is shown that the class \textit{features}\footnote{In this paper, unless specified, \textit{features} are assumed to be abstract representations derived from \textit{patterns} in the data distribution (e.g., how round the objects in an image are), rather than the set of individual \textit{attributes} that characterize the data (e.g., the set of pixels of an image).} associated to such directions are, indeed, the most relevant ones as far as the classification performance of the model is concerned, what links the accuracy of DNNs with their vulnerability to adversarial attacks.

The aforementioned theoretical frameworks focus, in particular, on the vulnerability to universal perturbations. In this paper, we focus instead on one particular property of universal perturbations: the existence of \textit{dominant} classes that are significantly more frequently predicted for the perturbed (and misclassified) inputs. This phenomenon was first reported in \cite{moosavi-dezfooli2017universal} for image classification tasks. Subsequent works have also reported the existence of dominant classes in image classification tasks \cite{hirano2020universal,co2020universal}, and in text classification tasks \cite{behjati2019universal}. In this paper, we show that this happens also for other domains, such as speech command classification tasks in the audio domain. Although it is hypothesized in \cite{moosavi-dezfooli2017universal} that a possible explanation for the \textit{dominant} classes is that they occupy a larger region in the decision space, it is left as an open research question. In this paper, we tackle this research question and test multiple hypotheses in the search for a deeper understanding of this phenomenon.

Outside the particular field of universal perturbations, multiple theoretical frameworks have been proposed for the explanation of adversarial examples. Whereas most of them focus on the properties of the DNNs \cite{szegedy2014intriguing,goodfellow2015explaining,tanay2016boundary,stutz2019disentangling}, other alternative explanations have also been proposed. In this paper, special attention is paid to the one introduced in \cite{ilyas2019adversarial}, in which adversarial examples are explained in terms of the \textit{robustness} of the features in the data. In particular, it it shown that datasets contain non-robust features which, although being highly discriminative (i.e., that the data is well described by these features), are uncorrelated with the ground-truth classes when they are perturbed by small (adversarial) perturbations. Thus, when a classifier learns to rely on such non-robust features to accurately classify the data, it becomes vulnerable to adversarial perturbations. The small robustness of such features to small perturbations also implies their lack of meaning for humans, which explains the imperceptibility of the attacks. In our paper (Section \ref{sec:class_properties}), we hypothesize that the higher sensitivity of the model to certain features might explain the existence of dominant classes.

\section{Proposed Framework}
\label{sec:framework}
Let us consider a machine learning model $f : X \rightarrow Y$, with $X\subseteq \mathbb{R}^d$ and $Y=\{y_1,\dots,y_k\}$, trained to classify inputs $x\in X$ coming from a data distribution $x \sim \mu$ among one of the $k$ possible classes in $Y$.
To formally describe \textit{dominant classes}, let us denote $p^v_j$ the probability of misclassifying an input as the class $y_j$ when a universal perturbation $v$ is added to the inputs:
\begin{equation}
\label{eq:p_dist}
    p^v_j =  \mathbb{P}_{\substack{x \sim \mu \\ f(x) \neq y_j}}\left[f(x+v)=y_j\right]
\end{equation}
Similarly, let $t^v_{i,j}$ represent the probability that, departing from an input of ground-truth $y_i$, the model incorrectly predicts the class $y_j$ for the perturbed inputs:
\begin{equation}
\label{eq:t_dist}
    t^v_{i,j} =  
    \mathbb{P}_{\substack{x \sim \mu \\ f(x)=y_i}}
    [f(x+v)=y_j]
\end{equation}
In practice, if the distribution $\mu$ is unknown, these probabilities can be estimated using a finite set of input samples $\mathcal{X}$.

\begin{definition}
$y_a$ is an attractor class for another class $y_i$ ($i \! \neq \! a$), under a perturbation $v$, which will be denoted as $y_i \xrightarrow{v} y_a$, if at least the $\alpha>\frac{1}{k-1}$ proportion of the inputs corresponding to the class $y_i$ are predicted as $y_a$ when they are perturbed with $v$:
\begin{equation}
t^v_{i,a} \geq \alpha.
\end{equation}
\end{definition}

\begin{definition}
$y_b$ is a dominant class for the universal perturbation $v$ if at least the $\beta  >  \frac{1}{k-1}$ proportion of the inputs are wrongly classified as $y_b$ when they are perturbed with $v$:
\begin{equation}
\label{eq:def_attractor}
p^v_b \geq \beta.
\end{equation}
\end{definition}
Alternatively, $y_b$ can be defined also in terms of the number of classes that it attracts. Precisely, $y_b$ is dominant if it is an attractor class for at least the $\zeta>\frac{1}{k-1}$ proportion of the remaining classes:
\begin{equation}
\frac{|Y_B|}{k-1}\geq\zeta, \textrm{ where } \ Y_B = \{ y_i \in Y \mid  y_i \xrightarrow{v} y_b \}.
\end{equation}
The choice of the parameters $\alpha$, $\beta$ and $\zeta$ can determine the existence of multiple attractor and dominant classes. In this paper, we assume $\alpha,\beta,\zeta \geq \frac{1}{3}$ since we are interested in those classes which are incorrectly predicted for a significant proportion of inputs, or which attract a significant proportion of other classes.

To explain the relationship between universal perturbations and dominant classes, we use a speech command classification problem in the audio domain as a testbed. We selected the Speech Command Dataset \cite{warden2018speech}, in which the underlying task consists of classifying audio signals, of fixed length, into one of the following classes: \textit{silence, unknown, yes, no, up, down, left, right, on, off, stop} and \textit{go}.  

We trained a convolutional neural network as a classifier, based on the architecture proposed in \cite{sainath2015convolutional}, which is composed of two convolutional layers with ReLU activations, a fully connected layer and a final softmax layer. This architecture has been used in a number of related works \cite{warden2018speech,li2020advpulse,alzantot2018did,yu2018robust}. The audio waveforms (in the time-domain) from the input space $\mathbb{R}^{16000}$, which take values in the range $[-1,1]$, are first converted into spectrograms by dividing the audios into frames of 20ms, with a stride of 10ms, and applying the real-valued fast Fourier transform (retrieving 512 components) for each frame. As the frequency spectrum of a real signal is Hermitian symmetric, only the first 257 components are retained. The dimension of the resulting spectrogram is $99  \times 257$. Finally, the Mel-Frequency Cepstrum Coefficients (MFCCs)  \cite{muda2010voice} are extracted from the spectrogram, in the space $\mathbb{R}^{99 \times 40}$, before being sent to the network. It is worth pointing out that the adversarial perturbations that are generated for this model are optimized in an end-to-end fashion, directly in the audio waveform representation of the signal. 

We selected the UAP-HC algorithm introduced in \cite{vadillo2019universal} to create the universal perturbations. This algorithm, which is a reformulation for the audio domain of the one proposed in \cite{moosavi-dezfooli2017universal}, consists of iteratively accumulating individual untargeted adversarial perturbations, generated using the DeepFool algorithm \cite{moosavi-dezfooli2016deepfool}. The pseudocodes for both the UAP-HC and DeepFool algorithms can be found in Algorithm \ref{alg:UAP-HC} and Algorithm \ref{alg:deepfool}, respectively. These algorithms have been generalized to (optionally) prevent them from reaching certain adversarial classes. This generalization will be further described and motivated in Section \ref{sec:dominant_classes_in_our_domain}.

Finally, we highlight that the rationale of the DeepFool algorithm relies on a geometric approach. In particular, a first-order approximation of the decision boundaries is used to move the input towards the estimated closest boundary, being, therefore, an untargeted attack. Thus, the optimization process of the UAP-HC algorithm is not biased towards any particular class, although, in practice, different universal perturbations lead in most of the cases to the same dominant classes.

\begin{algorithm}[]
 \caption{UAP-HC}
  \label{alg:UAP-HC}
\begin{algorithmic}[1]
 \REQUIRE { A classification model $f$,  a set of input samples $\mathcal{X}$, a projection operator $\mathcal{P}_{p,\xi}$, a fooling rate threshold $\delta$, a maximum number of iterations $I_\text{max}$, a set of restricted classes $\mathcal{R} \subset Y$ }
 \ENSURE{ A universal perturbation $v$}
 
 \STATE $v \gets $ initialize with zeros
 \STATE $FR \gets$ 0 \quad \algcomment{Fooling rate.}
 \STATE $iter \gets$ 0 \quad  \algcomment{Iteration number.}
 \WHILE{$FR < 1-\delta \wedge iter< I_\text{max} $}
  \STATE $\mathcal{X} \gets $ randomly shuffle $\mathcal{X}$
  \FOR{$x_i \in \mathcal{X}$}
  	  \STATE \algcomment{Check that $x_i$ is not already fooled by $v$:}
  	  \IF{$f(x_i+v) = f(x_i)$}
	  	\STATE $\bigtriangleup v_i \leftarrow$ DeepFool($x_i + v$, $f$, $\mathcal{R}$)\\
	  	\STATE $v' \gets \mathcal{P}_{p,\xi}(v+\bigtriangleup v_i)$ \quad \algcomment{ Project $(v+\bigtriangleup v_i)$ in the $\ell_p$ ball of radius $\xi$ and  centered at 0.} 
	  	\STATE $FR' \gets \displaystyle \mathbb{P}_{x\in \mathcal{X}}\left[ f(x)\neq f(x+v') \right]$
	  	\STATE \algcomment{Update $v$ only if adding $\bigtriangleup v_i$ increases the FR and if the current class is not in $\mathcal{R}$:}
	  	\IF{$FR < FR' \wedge f(x_i+v+\bigtriangleup v_i)\notin \mathcal{R}$}
			\STATE $v \gets v'$\\	 
  			\STATE $FR \gets FR'$
	  	\ENDIF	 
	 \ENDIF  
  \ENDFOR
  \STATE $iter \gets iter + 1 $\quad 
 \ENDWHILE
\end{algorithmic}
\end{algorithm}

\begin{algorithm}[]
 \caption{DeepFool}
  \label{alg:deepfool}
\begin{algorithmic}[1]
 \REQUIRE {An input sample $x$ of class $y_i$, a classifier $f$, a set of restricted classes $\mathcal{R}\subset Y$.}
 \ENSURE{ An individual perturbation $r$.}

 \STATE $x' \gets x$
 \STATE $r \ \gets $ initialize with zeros 
 \STATE $Y' \gets Y- \left( \mathcal{R}\cup \{ y_i \} \right)$
 \WHILE{$f(x') = y_i $}
    \FOR{$y_j \in Y'$}
 	    \STATE $f_j' \ \gets f_j(x') - f_i(x')$ 
  	    \STATE $w_j' \gets \bigtriangledown f_j(x') - \bigtriangledown f_i(x')$
  	\ENDFOR
  	\STATE $l \gets \argmin_{j\in Y'} \frac{|f'_j|}{||w'_j||}$
  	\STATE $r \gets r + \frac{|f'_l|}{||w'_l||_2^2}w'_l$  
	\STATE $x' \gets x + r$
 \ENDWHILE
\end{algorithmic}
\end{algorithm}

\section{Dominant classes in speech command classification}
\label{sec:dominant_classes_in_our_domain}

In this section, we generate different universal adversarial perturbation for the speech command classification task described in Section \ref{sec:framework}, in order to investigate whether in this domain dominant classes are also produced.

We start by generating 10 different universal perturbations using the UAP-HC algorithm, without restricting any class ($\mathcal{R}=\varnothing$). We set $\xi=0.1$ as threshold for the perturbation $\ell_2$ norm, and restricted the UAP-HC algorithm to a maximum of five iterations. To generate the perturbations, we used a \textit{training} set of 100 inputs per class, which makes a total of 1200 inputs. Once the perturbations are generated, their effectiveness will be measured in a \textit{test} set, containing samples that were not used during the generation of the perturbations. The initial accuracy of the model in this set is 85.52\%.\footnote{The number of samples per class in the test set and the accuracy of the model in each class is reported in Table \ref{tab:test_clean_acc}.}

\begin{figure}[]
    \centering
    \includegraphics[scale=0.31]{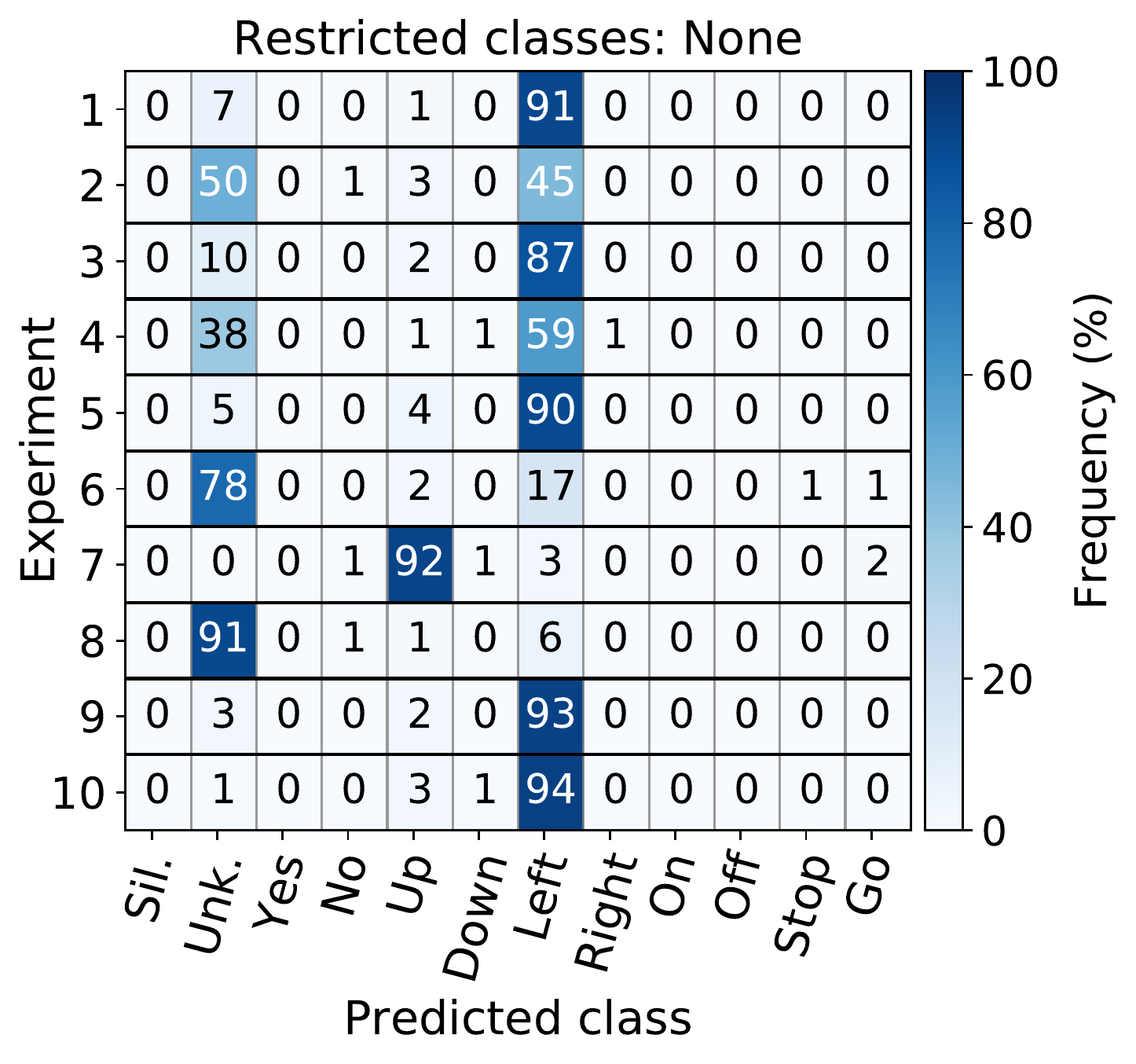}
    \includegraphics[scale=0.31]{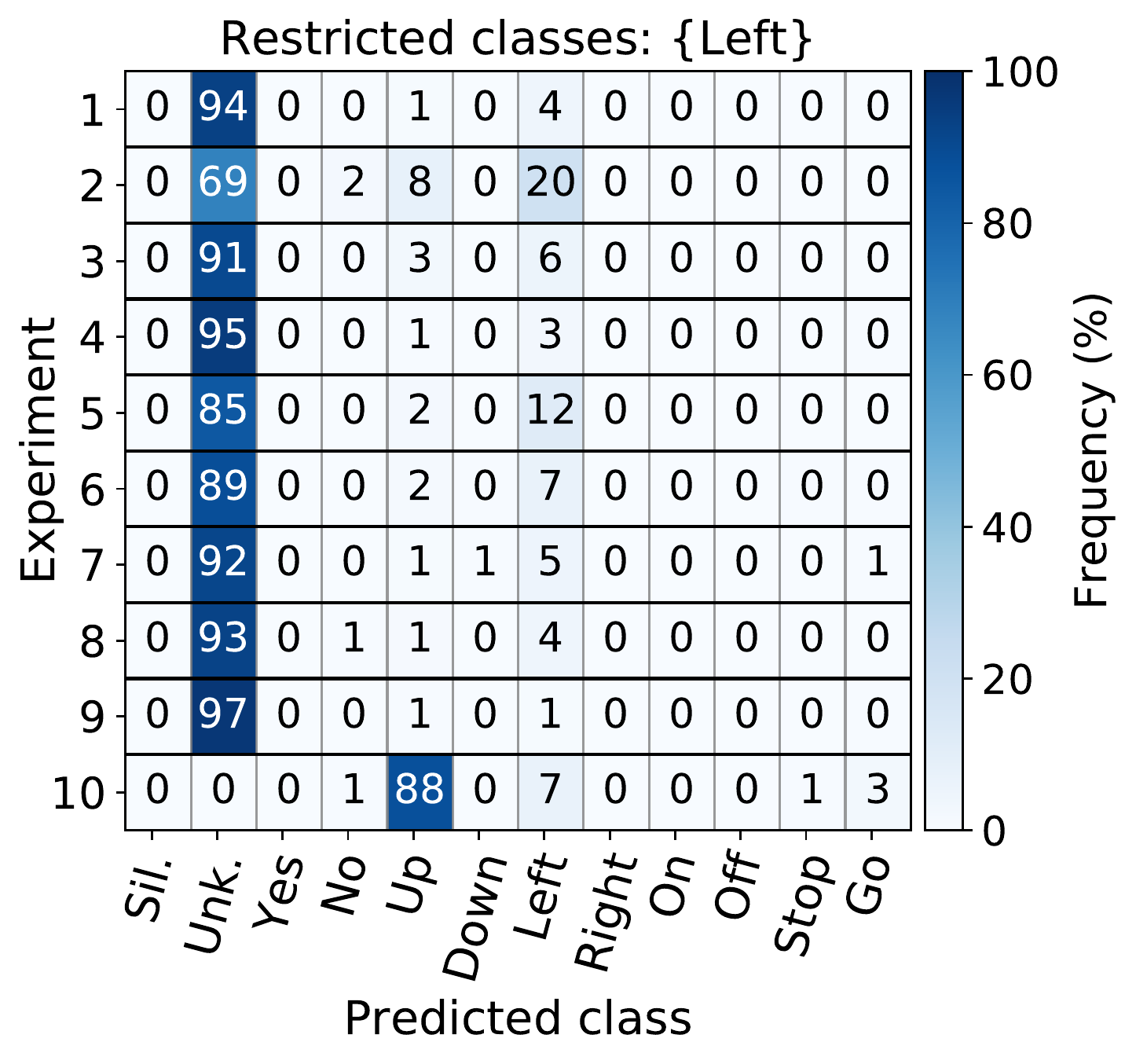}
    \includegraphics[scale=0.31]{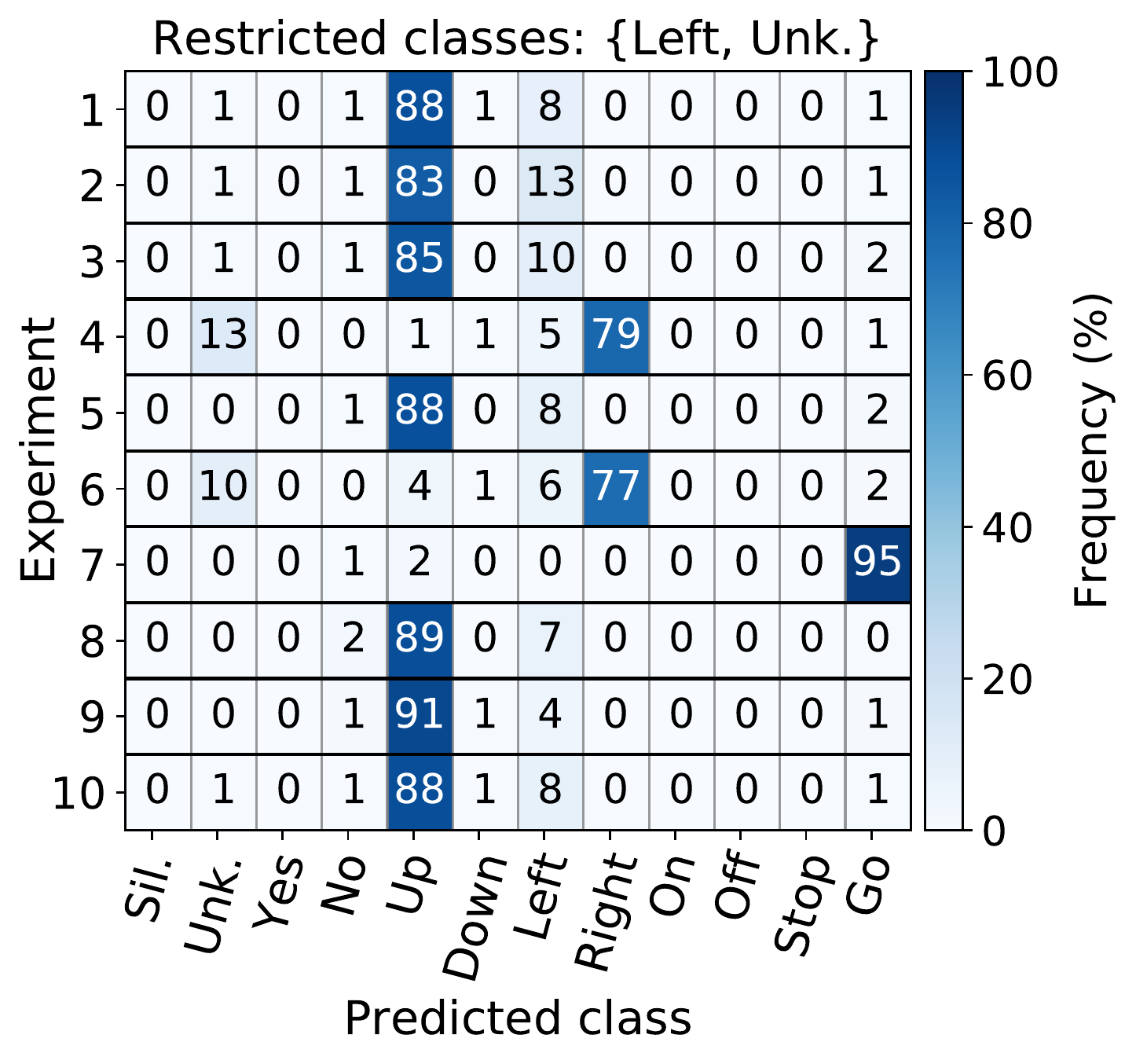}
    \caption{Overview of the frequency with which each class was assigned to the inputs misclassified as a consequence of universal perturbations. The frequencies have been computed individually (row-wise) for the 10 perturbations generated in each of the following configurations of the UAP-HC algorithm: default algorithm (left), restricting the algorithm to follow the class \textit{left} (center) and restricting the algorithm to follow the classes \textit{left} and \textit{unknown} (right).}
    \label{fig:univ_restricting_classes}
\end{figure}

According to the results, the algorithm led to universal perturbations with \textit{left} and \textit{unknown} as dominant classes for almost all the experiments. This can be seen in Figure \ref{fig:univ_restricting_classes} (left), which  shows the frequency with which each class is wrongly predicted when the perturbation is applied to the audios in the test set. We only considered those inputs that were initially correctly classified by the model, but misclassified when the perturbation is applied. The frequencies are shown individually for the ten universal perturbations, with each row corresponding to one perturbation. As can be seen, both \textit{left} and \textit{unknown} arise as dominant classes in 9 of the 10 experiments, sometimes even at the same time.

It is important to highlight that dominant classes arise without being imposed in the universal perturbation crafting procedure. However, we tested whether dominant classes remain dominant even if we explicitly avoid them during the optimization process (see Algorithms \ref{alg:UAP-HC} and  \ref{alg:deepfool}). We start by preventing the algorithm from considering those directions that point to the decision boundaries of the class \textit{left}. The results obtained for ten new perturbations generated with this restriction are shown in Figure \ref{fig:univ_restricting_classes} (center). As can be seen, the most frequent adversarial class is now \textit{unknown} for 9 of the 10 perturbations created. 

We went another step further and repeated the experiment, this time, however, restricting the boundaries corresponding to both \textit{left} and \textit{unknown} classes. The results are shown in Figure \ref{fig:univ_restricting_classes} (right). In this case, the two restricted classes were no longer dominant classes, but different dominant classes were obtained, precisely, \textit{up}, \textit{right} and \textit{go}. It is also worth emphasizing that, although dominant classes were obtained in all the experiments, they were different depending on which other classes were restricted. For instance, whereas the class \textit{up} rarely appeared as dominant without restrictions, it is the most frequent dominant class when both \textit{left} and \textit{unknown} classes are restricted. 

\begin{figure}[]
    \centering
    \includegraphics[scale=0.32]{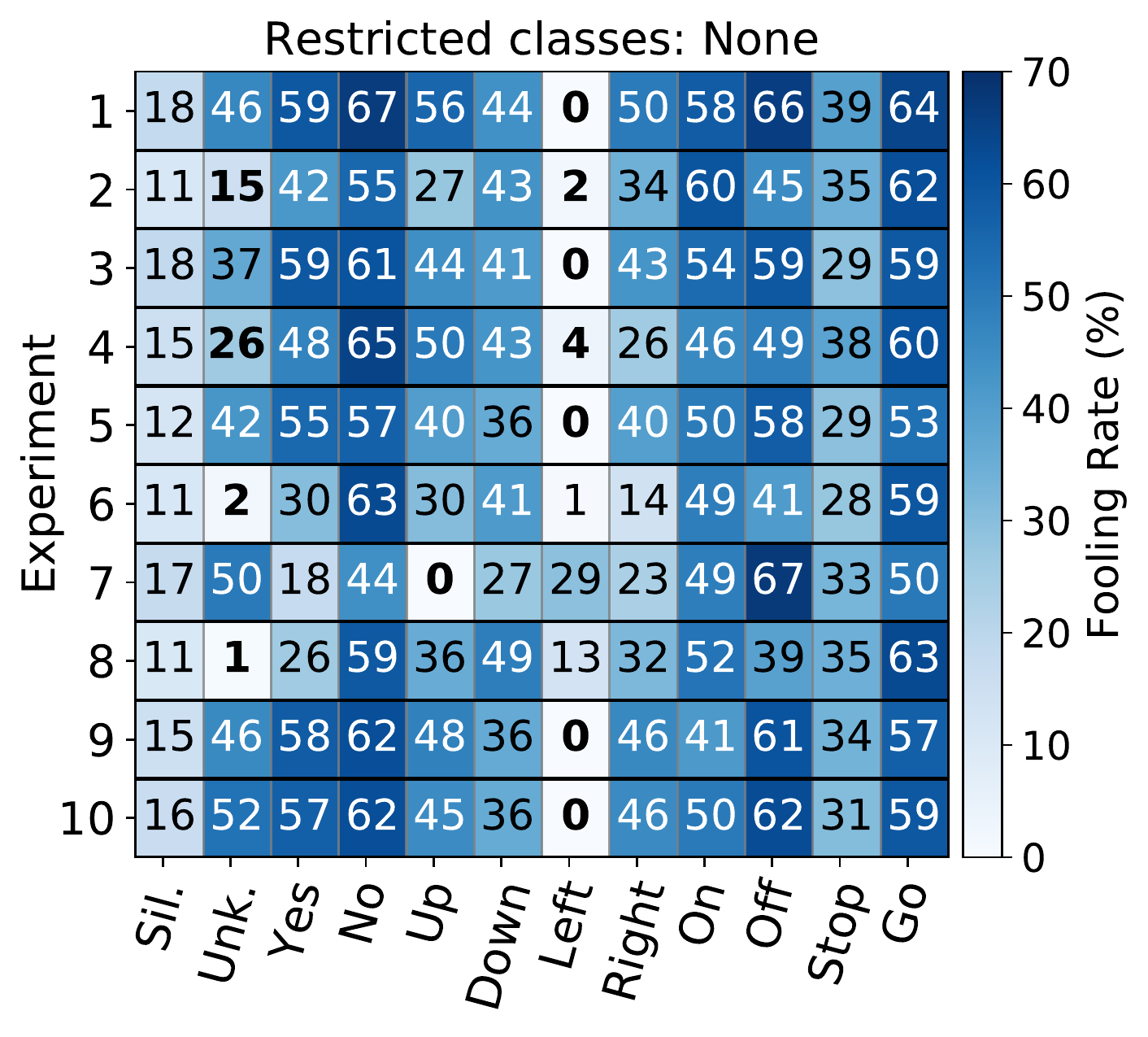}
    \includegraphics[scale=0.32]{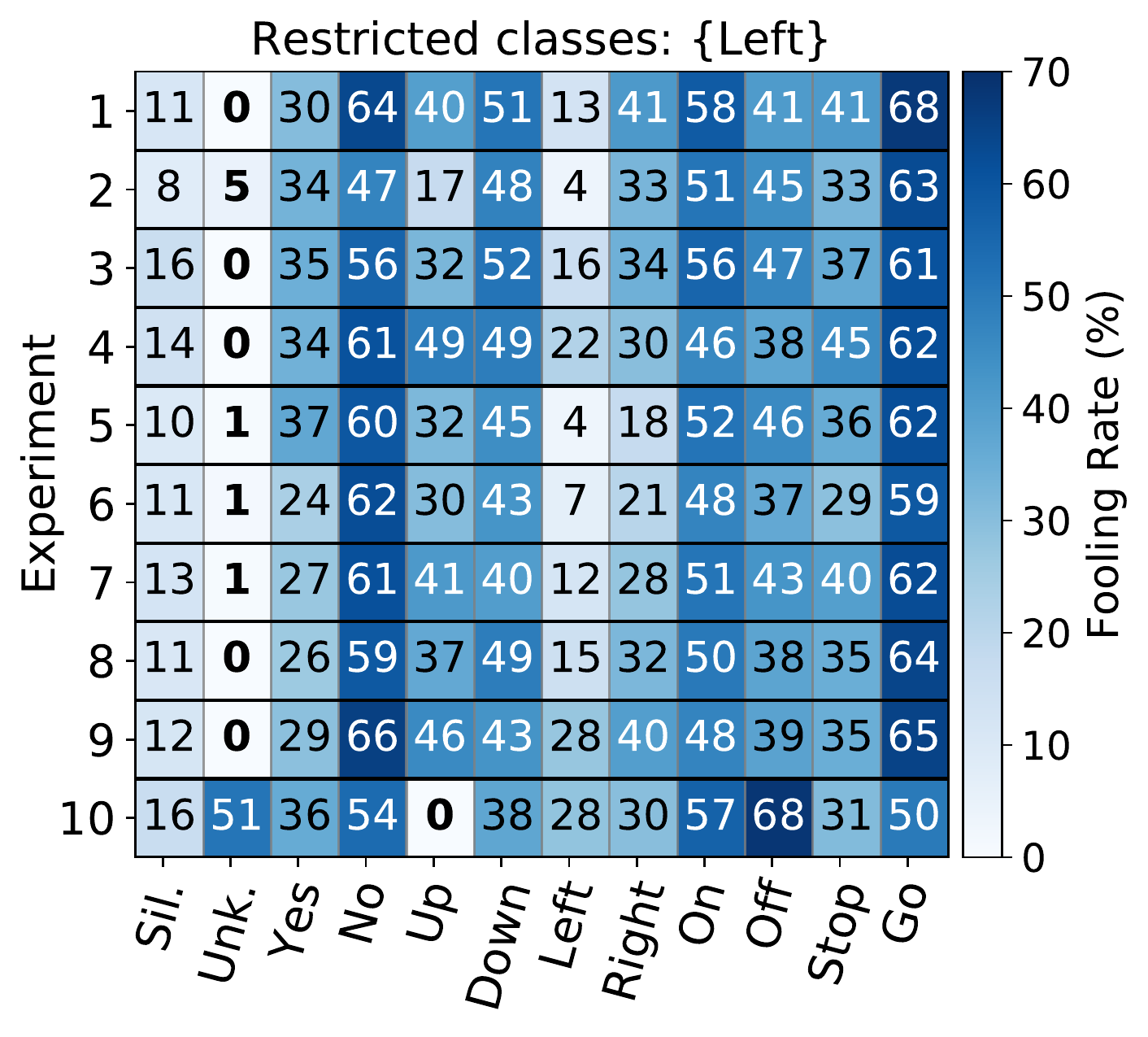}
    \includegraphics[scale=0.32]{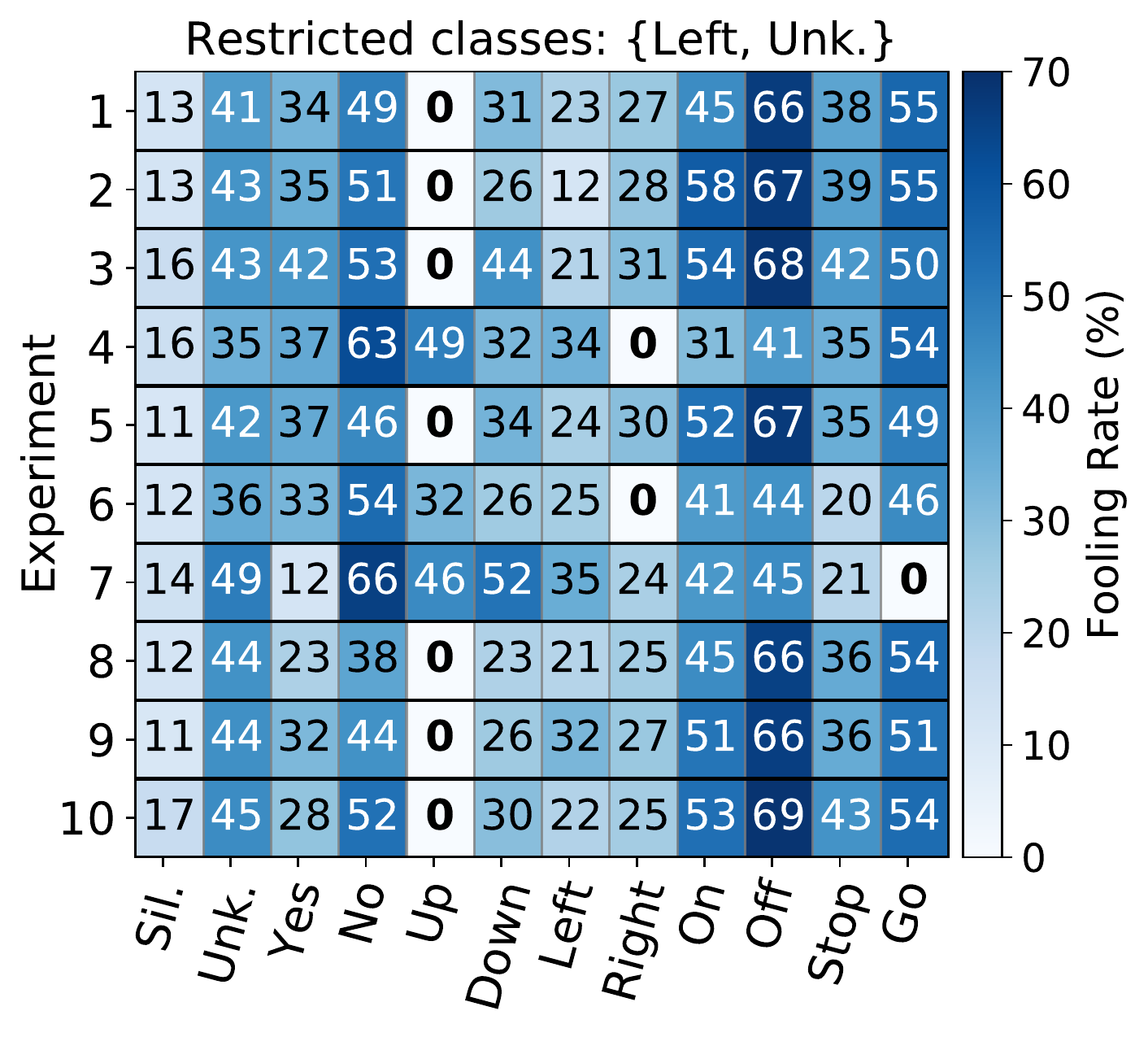}
    \caption{Fooling rate percentage, computed individually for each class, of the 10 perturbations generated in each of the following configurations of the UAP-HC algorithm: default algorithm (left), restricting the algorithm to follow the class \textit{left} (center) and restricting the algorithm to follow the classes \textit{left} and \textit{unknown} (right). In the three figures, the results corresponding to the dominant classes (for each experiment) have been highlighted using bold text.}
    \label{fig:univ_fr_classes}
\end{figure}

Regarding the effectiveness of the attacks, the fooling rate of every perturbation (i.e., the percentage of inputs that are misclassified when the perturbation is applied) is shown in Figure \ref{fig:univ_fr_classes}, for each class independently. The fooling rates have been computed considering the inputs that were initially correctly classified. As can be seen, the effectiveness of each perturbation is higher in some classes than in others, achieving up to $\approx$69\% in some cases. The fooling rates corresponding to the dominant classes, which have been highlighted in the figure, are practically zero for most of the perturbations, which reveals that the perturbation does not change the prediction of the model for those inputs.

For more informative results, the mean and maximum fooling rate of all the perturbations are shown in Table \ref{tab:univ_fr}. To avoid biases, these aggregated fooling rates have been computed in three different ways: (I) considering all the inputs, (II) without considering the inputs corresponding to the dominant classes, and (III) without considering the dominant classes and the class \textit{silence}. The reason for not considering the inputs belonging to the dominant classes is because the perturbation reinforces the confidence on those classes, and, as a consequence, there are practically no misclassifications in those inputs. On the contrary, the results for the class \textit{silence} are clearly lower than for the rest of the classes, which biases the results. Comparing the average effectiveness of the universal perturbations, we can notice that the average fooling rate achieved by the perturbations decreases when the dominant classes are restricted in the UAP-HC algorithm.


\begin{table}[!h]
\centering\sisetup{table-column-width=3cm}
\begin{tabular}{@{}lr@{\hskip 0.5in}rr@{\hskip 0.5in}rrr@{}}
\toprule
\multirow{3}{*}{\begin{tabular}[c]{@{}l@{}} \\ Restricted \\ classes in \\ UAP-HC\end{tabular}} & \multicolumn{6}{c}{Fooling Rate} \\ 
\cmidrule(l){2-7}
 & \multicolumn{2}{c}{\begin{tabular}[c]{@{}c@{}}  Considering all \\ the classes \end{tabular}} & \multicolumn{2}{c}{\begin{tabular}[c]{@{}c@{}}w/o considering\\ \ dominant classes \ \end{tabular}} & \multicolumn{2}{c}{\begin{tabular}[c]{@{}c@{}}w/o considering \\ dominant \& \textit{Silence} \end{tabular}} \\ 
 \cmidrule(l){2-3} \cmidrule(l){4-5} \cmidrule(l){6-7}
 & Mean & Max. & Mean & Max. & Mean & Max. \\ \midrule
None & 
37.94 & 46.34 & 41.68 & 50.84 & 44.97 & 54.76 \\
\{\textit{Left}\} & 
34.90 & 37.73 & 37.39 & 40.60 & 40.32 & 43.71 \\
\{\textit{Left}, \textit{Unk}.\} & 
33.75 & 37.49 & 37.08 & 41.36 & 39.90 & 44.37 \\
\bottomrule
\end{tabular}
\caption{Effectiveness of the UAP-HC algorithm in a set of \textit{test} samples, not seen during the generation of the perturbations.}
\label{tab:univ_fr}
\end{table}

\section{Hypotheses about the existence dominant classes}
\label{sec:hypotheses}

In this section, we propose a number of hypotheses to explain and characterize the relationship between universal adversarial perturbations and dominant classes. The proposed hypotheses are also experimentally tested using the framework described in Section \ref{sec:framework}.

\subsection{Dominant classes occupy a larger region in the input space}

In \cite{moosavi-dezfooli2017universal}, the existence of dominant classes is attributed to a larger region of such classes in the image space. Nevertheless, due to the high dimensionality of the input spaces in current machine learning problems, exploring the volume that each decision region occupies in the whole input space is intractable in practice.

Even so, to test this hypothesis, we randomly sampled and classified 1,000,000 inputs from the input space. The values of the inputs were sampled uniformly at random in the range $[-1,1]$.  We found that all the samples were classified as the class \textit{silence}, which is not a dominant class in our experiments, as shown in Section \ref{sec:dominant_classes_in_our_domain} (see Figure \ref{fig:univ_restricting_classes}). Therefore, our results suggest that there is not necessarily a connection between the \textit{volume} occupied by the decision regions of different classes and the frequency with which inputs perturbed by universal perturbations reach the regions corresponding to the dominant classes.

\subsection{Class properties of universal perturbations}
\label{sec:class_properties}
Universal perturbations are capable of changing the output class of a large number of inputs, and the majority of the misclassified inputs are moved unintentionally towards a dominant class. In this section, we show that the perturbation itself is predicted by the model as the dominant class with high confidence.

\begin{figure}[]
    \centering
    \includegraphics[scale=0.4]{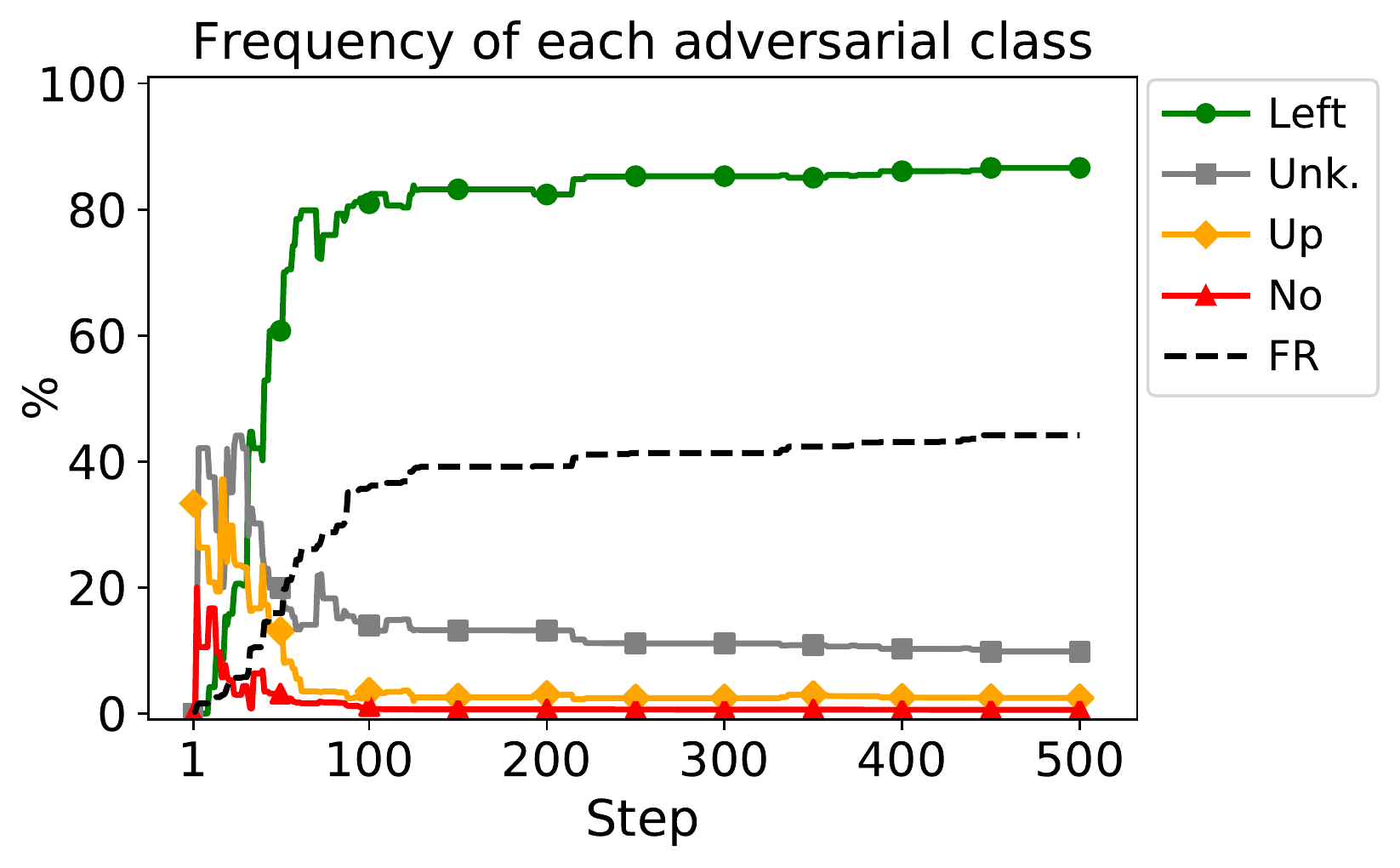}
    \includegraphics[scale=0.4]{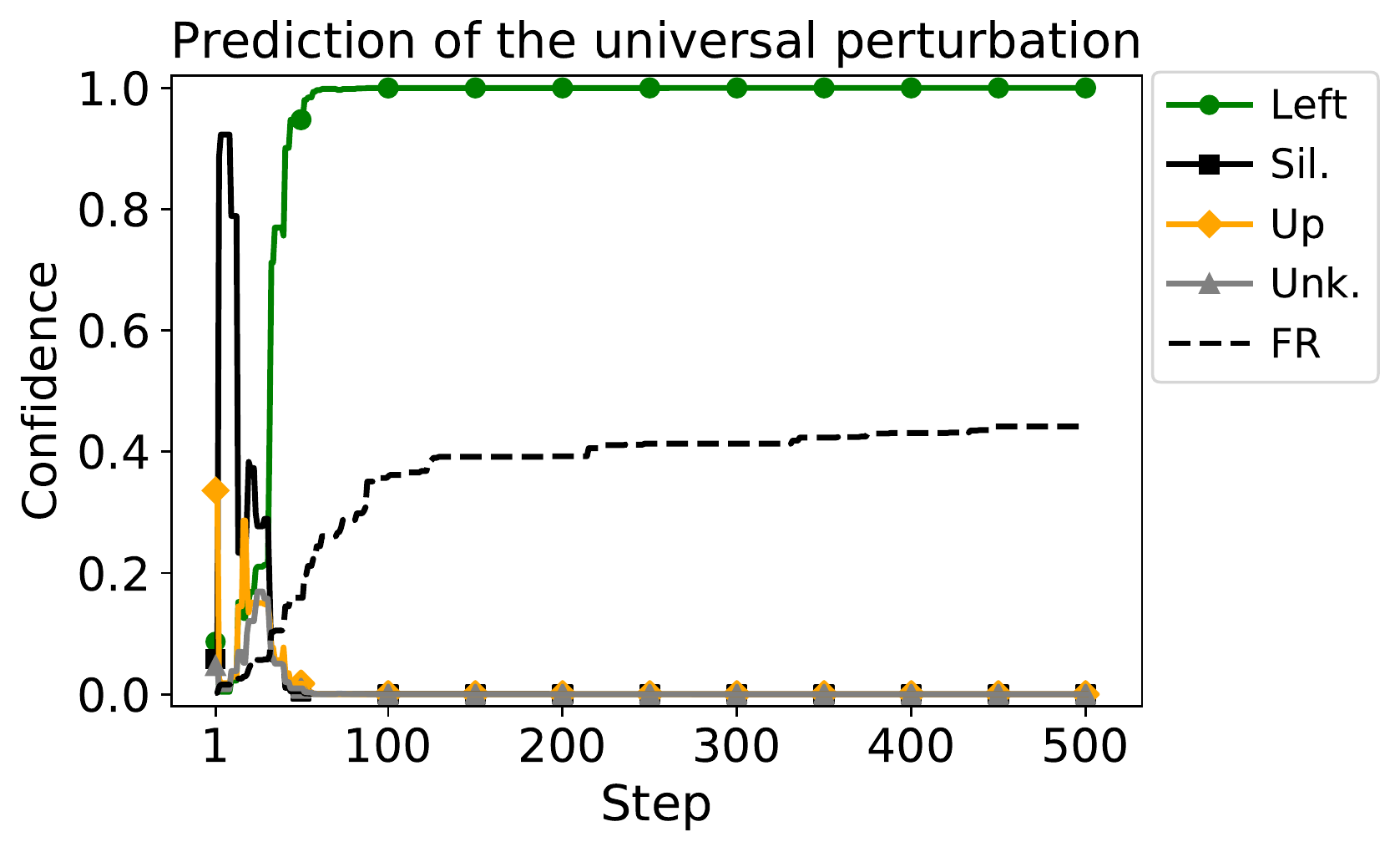}
    \caption{Evolution of three different factors during the optimization process of a universal perturbation using the UAP-HC algorithm: the frequency with which the inputs are classified as the dominant class (left), the confidence of the model in the dominant class when the perturbation is predicted (right), and the evolution of the fooling ratio (FR), which is shown in both plots as a reference. These results have been computed on the training set, and correspond to the first experiment reported in Section \ref{sec:dominant_classes_in_our_domain}, for the case in which no class was restricted. For the sake of clarity, only the information of the four most relevant classes are plotted in each plot.}
    \label{fig:univ_correlations}
\end{figure}

In fact, we noticed that the following three factors are positively correlated during the generation process of a universal perturbation $v$: the fooling rate ($\mathcal{F}_{1}$), the percentage of inputs misclassified as the dominant class $y_b$ ($\mathcal{F}_{2}$), and the confidence with which the model considers that the perturbation belongs to the dominant class ($\mathcal{F}_{3}$):\footnote{For those perturbations in which there are two dominant classes at the same time, the class $f(v)$ has been considered as the dominant (i.e., the class assigned to the perturbation by the model).}
\begin{align}
\mathcal{F}_{1}(v) & = \mathbb{P}_{x\in \mathcal{X}}\left[ f(x)\neq f(x+v) \right], \\
\mathcal{F}_{2}(v) & = \mathbb{P}_{x\in \mathcal{X}}\left[f(x)=y_b \right], \\
\mathcal{F}_{3}(v) & = f_b(v),
\end{align}
where $\mathcal{X}$ is a set of inputs and $f_j: X \rightarrow \mathbb{R}$ represents the output confidence of the classifier $f$ corresponding to the class $y_j$. An example of the evolution of these factors during the optimization process of a universal perturbation, using the UAP-HC algorithm, is shown in Figure \ref{fig:univ_correlations}. These results correspond to the first experiment of Section \ref{sec:dominant_classes_in_our_domain}, for the case in which no class was restricted. In particular, the left figure shows the evolution of the frequency with which each class is (wrongly) predicted for the misclassified inputs, and the right figure shows the output confidences of the model when the universal perturbation is classified. The fooling ratio of the perturbation has been included in both figures as a reference, represented by a dashed line.

For the 10 different universal perturbations generated in Section \ref{sec:dominant_classes_in_our_domain} (without restricting any class), the average Pearson correlation coefficient between $\mathcal{F}_{1}$ and $\mathcal{F}_{3}$ during the first iteration of Algorithm \ref{alg:UAP-HC} is $0.79$. Similarly, the average correlation between $\mathcal{F}_{1}$ and $\mathcal{F}_{2}$ is $0.87$, and the average correlation between $\mathcal{F}_{2}$ and $\mathcal{F}_{3}$ is $0.91$.

Motivated by this finding, we studied whether any perturbation $v$ that is classified by the model as one particular class with high confidence is capable of producing the same effect as a universal perturbation, that is, to force the misclassification of a large number of inputs by pushing them to the class $f(v)$. For this purpose, we defined the following optimization problem, in which the objective is to find a perturbation $v$, with a constrained norm, that maximizes the confidence of the model in one particular class $y_t$, $f_t(v)$, that is:
\begin{equation}
\label{eq:lp_ball_search}
\max_v \ \  f_t(v) \quad \textrm{s.t.} \ \  ||v||_2 \leq \xi .
\end{equation}
We launched 100 trials for each possible target class, starting from random perturbations.\footnote{The initial perturbations were randomly sampled from the input space $\mathbb{R}^{16000}$, where each value was sampled uniformly at random in the range $[-10^{-3}
,10^{-3}]$.} We used a gradient descent approach to optimize the perturbation, restricting the search to 100 gradient descent iterations, and setting a threshold of $\xi=0.1$ for the perturbation norm.

\begin{figure}
    \centering
    \includegraphics[align=c,scale  =0.44]{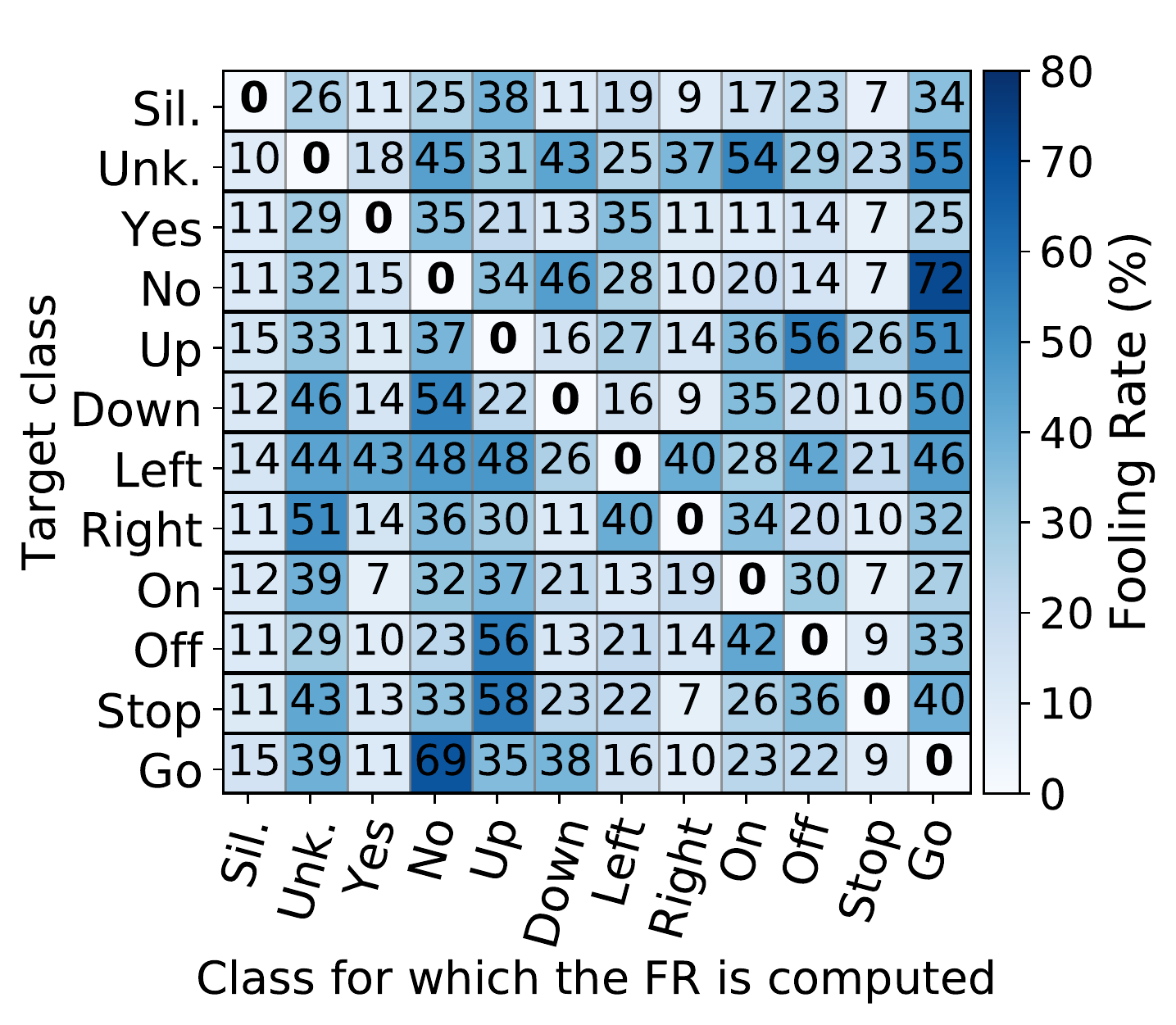}
    \hspace{0.2cm}
    \includegraphics[align=c,scale=0.44]{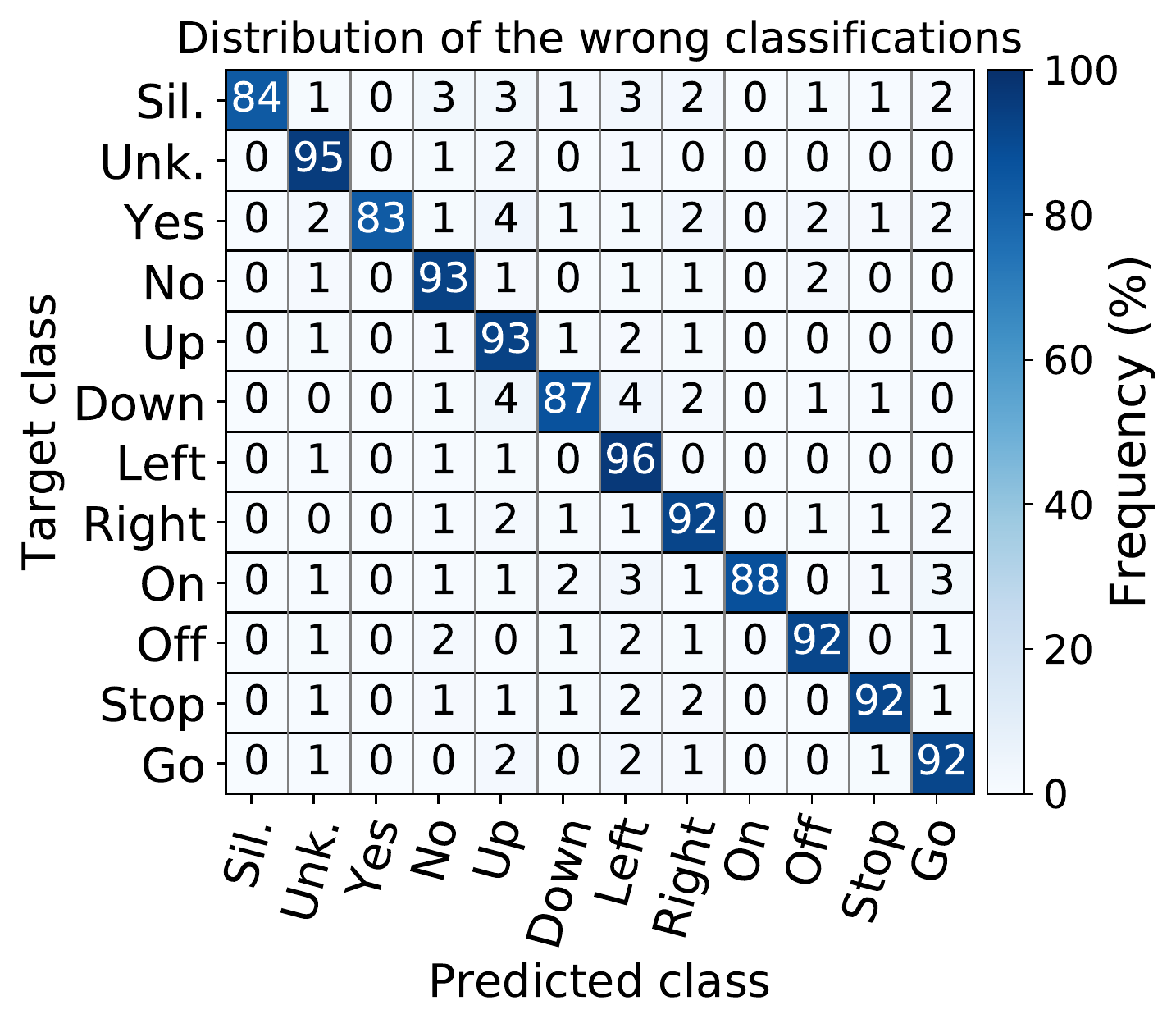}
    \caption{Overview of the effectiveness of the perturbations found by solving the optimization problem defined in \eqref{eq:lp_ball_search}. In both figures, the results are reported independently for each target class (row-wise), and are averaged for the 100 trials generated for each target class. Left: average fooling rate obtained by the 100 perturbations found for each target class, computed for each class individually. Right: Average frequency with which each class is wrongly assigned to the fooled inputs by the model.}
    \label{fig:lp_ball_results}
\end{figure}

The mean and maximum fooling rates obtained with the generated perturbations are shown in Table \ref{tab:lp_ball_fr}, computed independently for each target class. The fooling rate for each class individually is shown in Figure \ref{fig:lp_ball_results} (left). As can be seen in Table \ref{tab:lp_ball_fr}, the classes \textit{left} and \textit{unknown}, both the most frequent dominant classes associated to the universal perturbations generated using the UAP-HC algorithm (see Figure \ref{fig:univ_restricting_classes}), achieve a significantly higher effectiveness than the rest of classes. Apart from that, with independence of the target class, the majority of the samples fooled by these perturbations were classified as the target class. This is shown in Figure \ref{fig:lp_ball_results} (right), in which the average frequency with which each class is predicted under the effect of the perturbations is computed, independently for each target class.

Based on these results, we can hypothesize that the model is more sensitive to some class \textit{features} than to others, and that, ultimately, the sensitivity degree to each class feature is what determines the dominant classes. In other words, a class $y_j$ will have a greater dominance the more sensitive the model is to the patterns in the data distribution that are associated to $y_j$ (by the model itself).\footnote{These results could be related to the non-robust data-feature framework introduced in \cite{ilyas2019adversarial} or to the link between the class-identity associations of the model and the most \textit{vulnerable} directions in the input space studied in \cite{jetley2018friends} (see Section \ref{sec:related_work} for more details).}

\begin{table}[!h]
\centering\sisetup{table-column-width=3cm}
\begin{tabular}{@{}l@{\hskip 0.3in}r@{\hskip 0.5in}rr@{\hskip 0.5in}rrr@{}}
\toprule
\multirow{3}{*}{\begin{tabular}[c]{@{}l@{}} \\[6pt] Target \\ class\end{tabular}} & \multicolumn{6}{c}{Fooling Rate} \\ 
\cmidrule(l){2-7}
 & \multicolumn{2}{c}{\begin{tabular}[c]{@{}c@{}}  Considering all \\ the classes \end{tabular}} & \multicolumn{2}{c}{\begin{tabular}[c]{@{}c@{}}w/o considering\\ \ dominant classes \ \end{tabular}} & \multicolumn{2}{c}{\begin{tabular}[c]{@{}c@{}}w/o considering \\ dominant \& \textit{Silence} \end{tabular}} \\ 
 \cmidrule(l){2-3} \cmidrule(l){4-5} \cmidrule(l){6-7}
 & Mean & Max. & Mean & Max. & Mean & Max. \\ \midrule
\textit{Sil.} & 17.85 & 21.71 & 19.77 & 24.05 & 19.77 & 24.05 \\
\textit{Unk.} & 30.31 & 33.88 & 32.40 & 36.21 & 35.00 & 39.14 \\
\textit{Yes} & 16.91 & 20.40 & 18.67 & 22.52 & 19.59 & 23.89 \\
\textit{No} & 23.46 & 25.82 & 25.28 & 27.84 & 26.91 & 29.74 \\
\textit{Up} & 25.53 & 28.19 & 28.16 & 31.10 & 29.79 & 32.97 \\
\textit{Down} & 22.56 & 24.68 & 24.45 & 26.75 & 25.95 & 28.28 \\
\textit{Left} & 32.57 & 37.25 & 35.73 & 40.87 & 38.37 & 44.22 \\
\textit{Right} & 23.25 & 27.28 & 25.38 & 29.78 & 27.07 & 31.88 \\
\textit{On} & 19.50 & 22.43 & 21.25 & 24.45 & 22.40 & 25.94 \\
\textit{Off} & 21.56 & 24.46 & 23.39 & 26.54 & 24.83 & 28.48 \\
\textit{Stop} & 25.07 & 27.21 & 27.61 & 29.97 & 29.64 & 32.32 \\
\textit{Go} & 22.99 & 25.66 & 24.84 & 27.72 & 26.03 & 29.24 \\
\bottomrule
\end{tabular}
\caption{Effectiveness of the perturbations generated using Algorithm \ref{eq:lp_ball_search}, averaged for the 100 perturbations generated for each target class.}
\label{tab:lp_ball_fr}
\end{table}

\subsection{Singular Value Decomposition}
\label{sec:svd}
In \cite{moosavi-dezfooli2017universal}, the existence of universal perturbations for image classification DNNs is attributed, in part, to similar patterns in the geometry of decision boundaries around different points of the decision space. In particular, as described in Section \ref{sec:related_work}, perturbations \textit{normal} to the decision boundaries in the vicinity of natural inputs approximately span a very low-dimensional subspace, revealing that \emph{similar} perturbations are capable of changing the output class of different input samples. This was assessed experimentally for state-of-the-art DNNs, by computing the Singular Value Decomposition (SVD) of a matrix collecting normalized individual untargeted perturbations generated using the DeepFool algorithm.  Their results show that the decay of the singular values was considerably faster in comparison to the decay obtained from the decomposition of random perturbations (sampled from the unit sphere). This implies that the subspace spanned just by the first $d'\ll d$ singular vectors (i.e., those corresponding to the highest singular values) contained vectors normal to the decision boundaries in the vicinity of natural samples. Indeed, random perturbations sampled from such a subspace were capable of achieving a fooling rate of nearly 38\% on unseen inputs, whereas random perturbations (of the same norm) in the input space only achieved a fooling rate of approximately 10\% \cite{moosavi-dezfooli2017universal}.

In this section, we take this approach as a framework to study the existence of dominant classes. First, we will replicate the previous experiment to assess whether, in the audio domain, it is also possible to find a low-dimensional subspace of the input space collecting vectors normal to the decision boundaries of DNNs. Nevertheless, due to the input transformation process required to convert the raw audio signal into the MFCC representation (see Section \ref{sec:framework}), the results might differ depending on the data representation in which the analysis is done. Thus, we computed the SVD for a set of individual perturbations and different sets of random perturbations, under the three main representations for audio signals: raw audio waveform, spectrogram and MFCC coefficients.

\subsubsection{Analysis of the SVD of audio perturbations}
Let us consider a set of $n$ natural input samples $\mathcal{X}=\{x_1,\dots,x_n\}$. The individual perturbations were generated using the DeepFool algorithm, in the raw audio waveform representation:
\begin{align}
\label{eq:svd_indiv}
    \mathcal{V} = \left\lbrace  v_i  \mid v_i=\text{DeepFool}(x_i)  , \ i=1,\dots,n \right\rbrace .
\end{align}
The perturbations that these raw waveforms produce in both the spectrogram and MFCC representations are computed as $v_i'=g(x_i+v_i)-g(x_i)$, being $g$ the input transform function, which maps the raw audio waveforms into either a spectrogram  or the MFCC features:
\begin{equation}
\label{eq:svd_indiv_spec}
    \mathcal{V}_\text{SPEC} \ \ =  \left\lbrace  v_i^\text{spec} \mid v_i^\text{spec} \ = \ g_\text{SPEC}(x_i+v_i)  - \ g_\text{SPEC}(x_i) , \ i=1,\dots,n   \right\rbrace , 
\end{equation}
\begin{equation}
\label{eq:svd_indiv_mfcc}
    \mathcal{V}_\text{MFCC} \ =  \left\lbrace  v_i^\text{mfcc} \mid v_i^\text{mfcc}= g_\text{MFCC}(x_i+v_i) -  g_\text{MFCC}(x_i) , \ i=1,\dots,n \right\rbrace .
\end{equation}
The random perturbations were sampled uniformly at random from the raw input space:
\begin{equation}
    \mathcal{R} =  \left\lbrace r_i \mid \ r_i \in \mathbb{R}^{16000} \wedge r_i^1,\dots,r_i^d \sim \mathcal{U}(-1,1)  , \ i=1,\dots,n \right\rbrace .
\end{equation}
As in the case of adversarial perturbations, the corresponding perturbations in the frequency-domain representation are computed as:
\begin{equation}
\label{eq:rand_spec}
    \mathcal{R}_\text{SPEC} \  =  \left\lbrace  r_i^\text{spec} \mid r_i^\text{spec} \ =\ \ g_\text{SPEC}(x_i+r_i) - \ g_\text{SPEC}(x_i)  , \ i=1,\dots,n \right\rbrace , 
\end{equation}
\begin{equation}
\label{eq:rand_mfcc}
    \mathcal{R}_\text{MFCC}  =  \left\lbrace  r_i^\text{mfcc} \mid  r_i^\text{mfcc}= \ g_\text{MFCC}(x_i+r_i) -  g_\text{MFCC}(x_i)  , \ i=1,\dots,n \right\rbrace .
\end{equation}
In this case, the random perturbations were scaled to have a fixed $\ell_2$ norm of 0.1  before being applied to the inputs in Equations \eqref{eq:rand_spec} and \eqref{eq:rand_mfcc}.

Finally, for a more representative analysis,  we considered two additional sets of random perturbations, sampled uniformly at random from the space of spectrograms ($\mathbb{R}^{99\times 257}$) and the space of MFCC coefficients ($\mathbb{R}^{99\times 40}$):
\begin{equation}
\label{eq:svd_rand_unif_spec}
    \mathfrak{R}_\text{SPEC} \ =  \left\lbrace \mathfrak{r}_i \mid \mathfrak{r}_i \in \mathbb{R}^{99 \times 257} \wedge \mathfrak{r}_i^1 , \dots , \mathfrak{r}_i^d \sim \mathcal{U}(-1,1)  , \ i=1,\dots,n \right\rbrace
\end{equation}
\begin{equation}
\label{eq:svd_rand_unif_mfcc}   
    \mathfrak{R}_\text{MFCC} =  \left\lbrace \mathfrak{r}_i \mid \mathfrak{r}_i \in \mathbb{R}^{99 \times 40} \ \  \wedge \mathfrak{r}_i^1 , \dots , \mathfrak{r}_i^d \sim \mathcal{U}(-1,1)  , \ i=1,\dots,n \right\rbrace
\end{equation}
All the perturbations described in Equations \eqref{eq:svd_indiv}-\eqref{eq:svd_rand_unif_mfcc} were normalized before computing the SVD.

Figure \ref{fig:singular_values} compares the decay of the singular values (sorted in decreasing order), for all the sets of perturbations considered in Equations \eqref{eq:svd_indiv}-\eqref{eq:svd_rand_unif_mfcc}. The results corresponding to the raw waveform, spectrogram and MFCC representations are shown in the first, second and third row of the figure, respectively. Whereas the left column shows the singular values obtained with the SVD for each data representation, in the right column the decays are characterized by fitting exponential curves (depicted as dashed lines) with the following form:
\begin{equation}
y=\rho \cdot e^{-x\lambda}+\omega \ \ \ , \ \ \ \rho,\lambda,\omega\in\mathbb{R}.     
\end{equation}
A higher value of the decay factor $\lambda$ represents a faster decay. Note that the singular values have been scaled in the range $[0,1]$ before fitting the exponential curves, for a more uniform comparison.

Regarding the results in the raw waveform representation, the decay of the singular values is mainly linear for both individual and random perturbations, showing indeed a very similar decay in both cases. This means that there is not a set of singular vectors that is significantly more \textit{informative} than the rest, and, as a consequence, a large set of vectors would be needed to provide an approximate basis for the perturbations. Thus,  the perturbations do not show significant correlations in this representation.  Regarding the frequency-domain representations, the decays of the singular values corresponding to the perturbations sampled uniformly at random in the space of spectrograms ($\mathfrak{R}_\text{SPEC}$) and in the space of MFCC coefficients ($\mathfrak{R}_\text{MFCC}$) are also clearly linear.

However, considering the perturbations in the frequency domain produced by the raw waveform perturbations, either random or adversarial, the singular values decay exponentially. These results indicate, first, that even if the perturbations are generated in the raw audio waveform representation, it is necessary to go to the frequency-domain to observe informative patterns. This might be a fundamental difference between the image domain and the audio domain, as most of the analyses done in the former can be done directly in the \textit{raw} image space.  Secondly, the effect of audio perturbations in the frequency-domain can be characterized by just a small (in comparison to the dimensionality of the corresponding spaces) number of singular vectors. For instance, for the MFCC representation, the most relevant information is captured in less than the $\sim$150 first singular vectors (that is, those corresponding to the highest singular values).
The fact that this happens for both random or adversarial perturbations could imply, however, that the captured correlations are uninformative about the geometry of the decision boundaries around natural inputs, or, alternatively, about the vulnerability of the network to adversarial attacks. Nevertheless, in the reminder of this section we show that the SVD of individual adversarial perturbations not only provides a representative basis for input-agnostic perturbations, but also that this basis is strongly connected with the dominant classes. For the previous reasons, the rest of the analysis will focus on the MFCC feature space.

\newcommand\tmpsc{0.56}
\begin{figure}
    \centering
    \includegraphics[scale=\tmpsc]{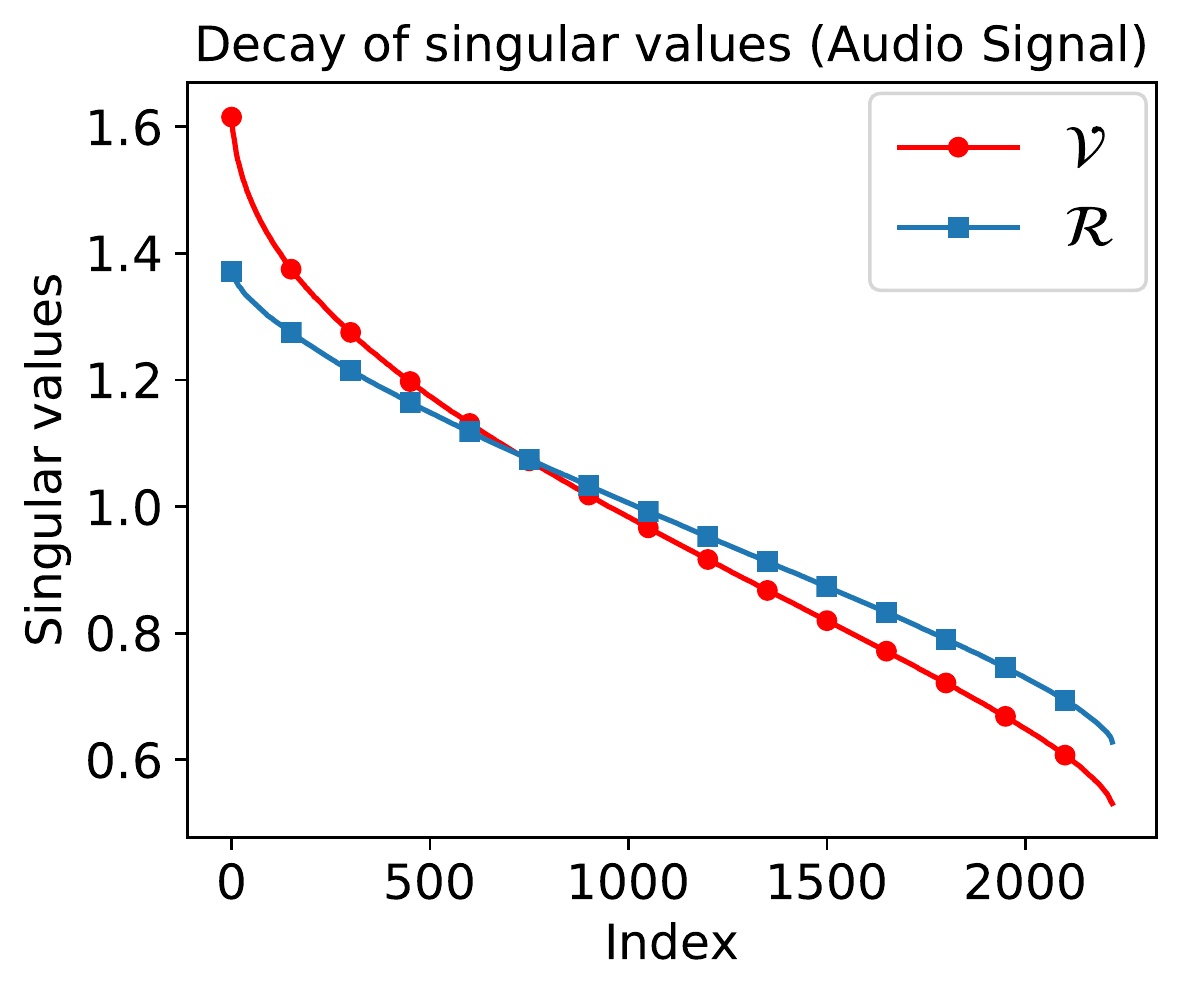}
    \includegraphics[scale=\tmpsc]{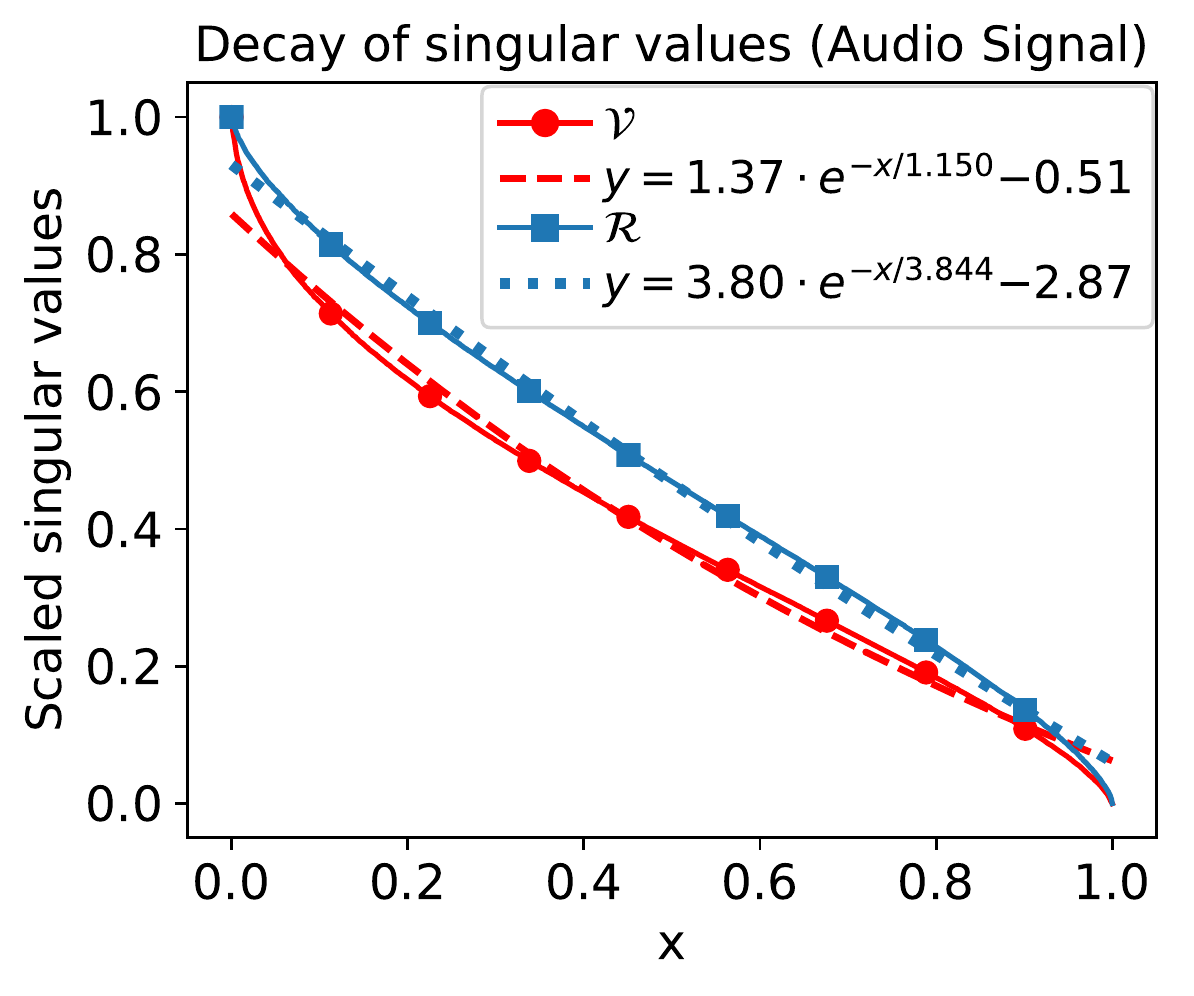}
    \includegraphics[scale=\tmpsc]{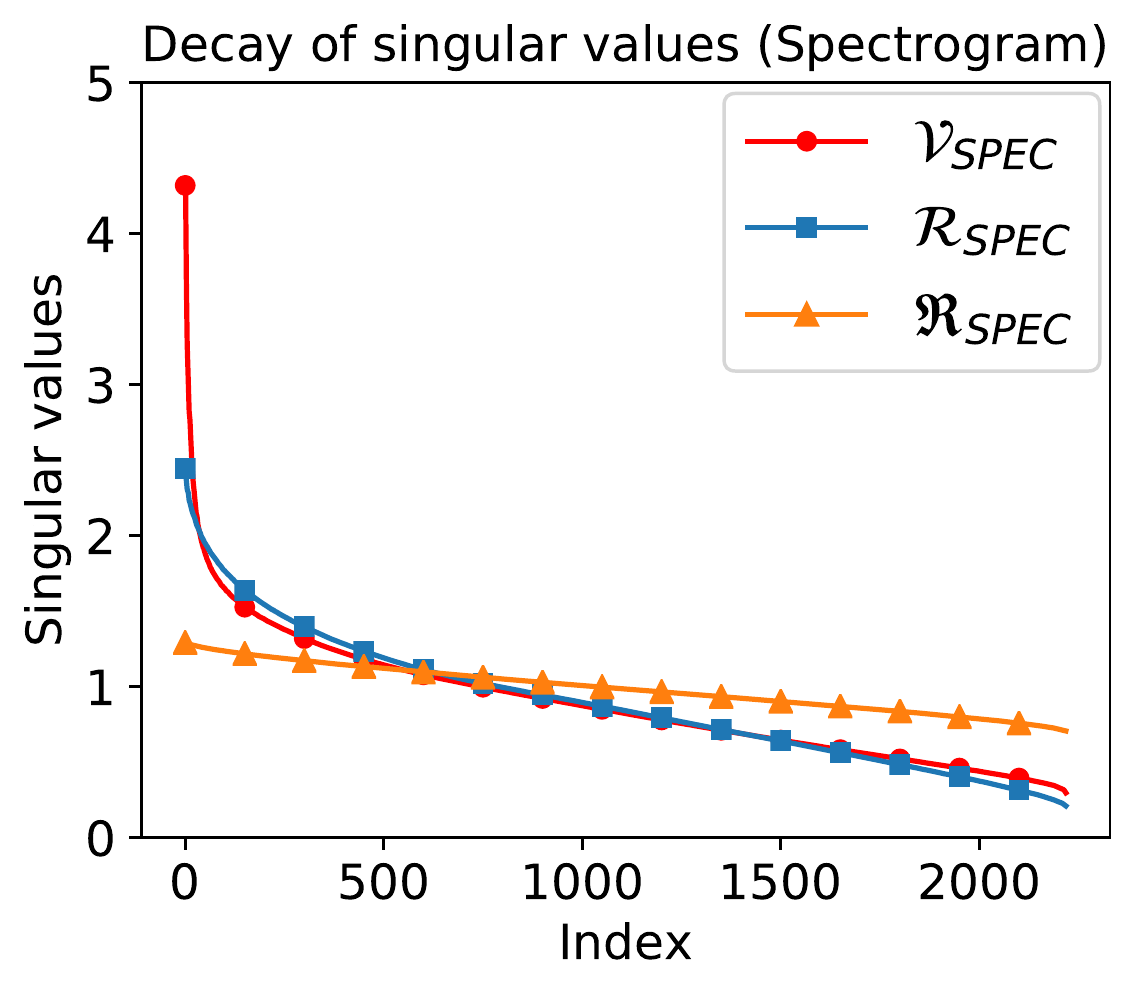}
    \includegraphics[scale=\tmpsc]{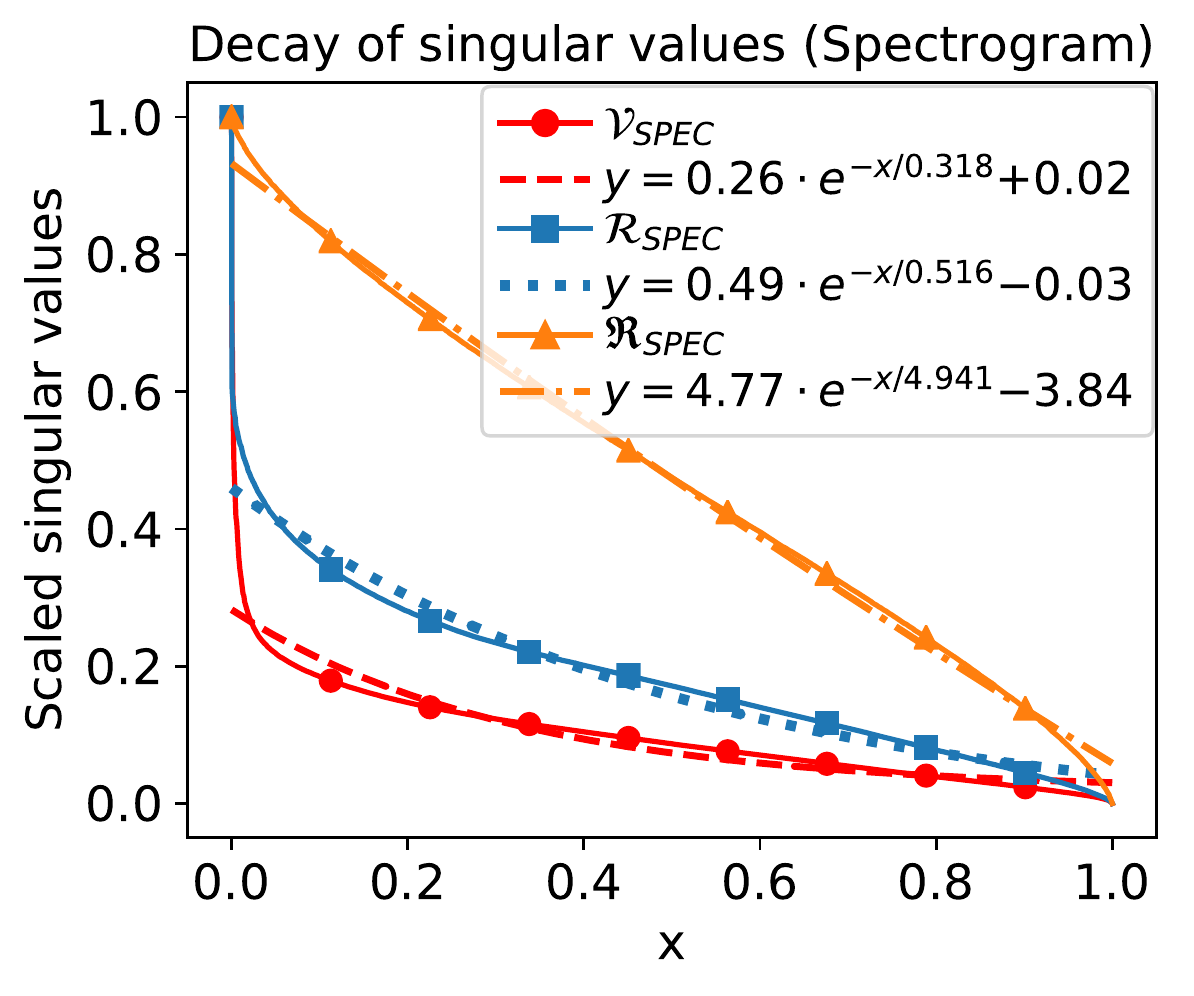}
    \includegraphics[scale=\tmpsc]{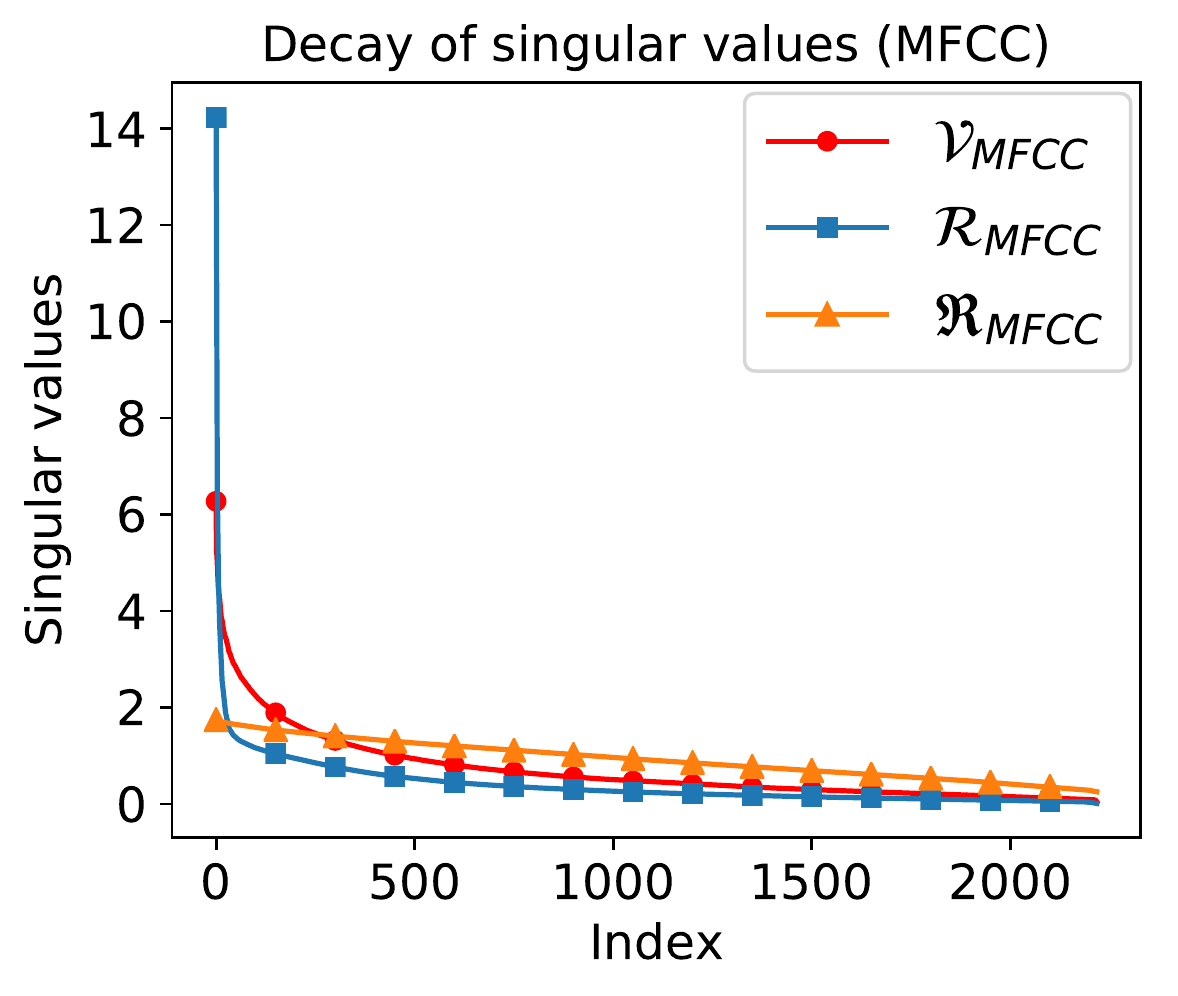}
    \includegraphics[scale=\tmpsc]{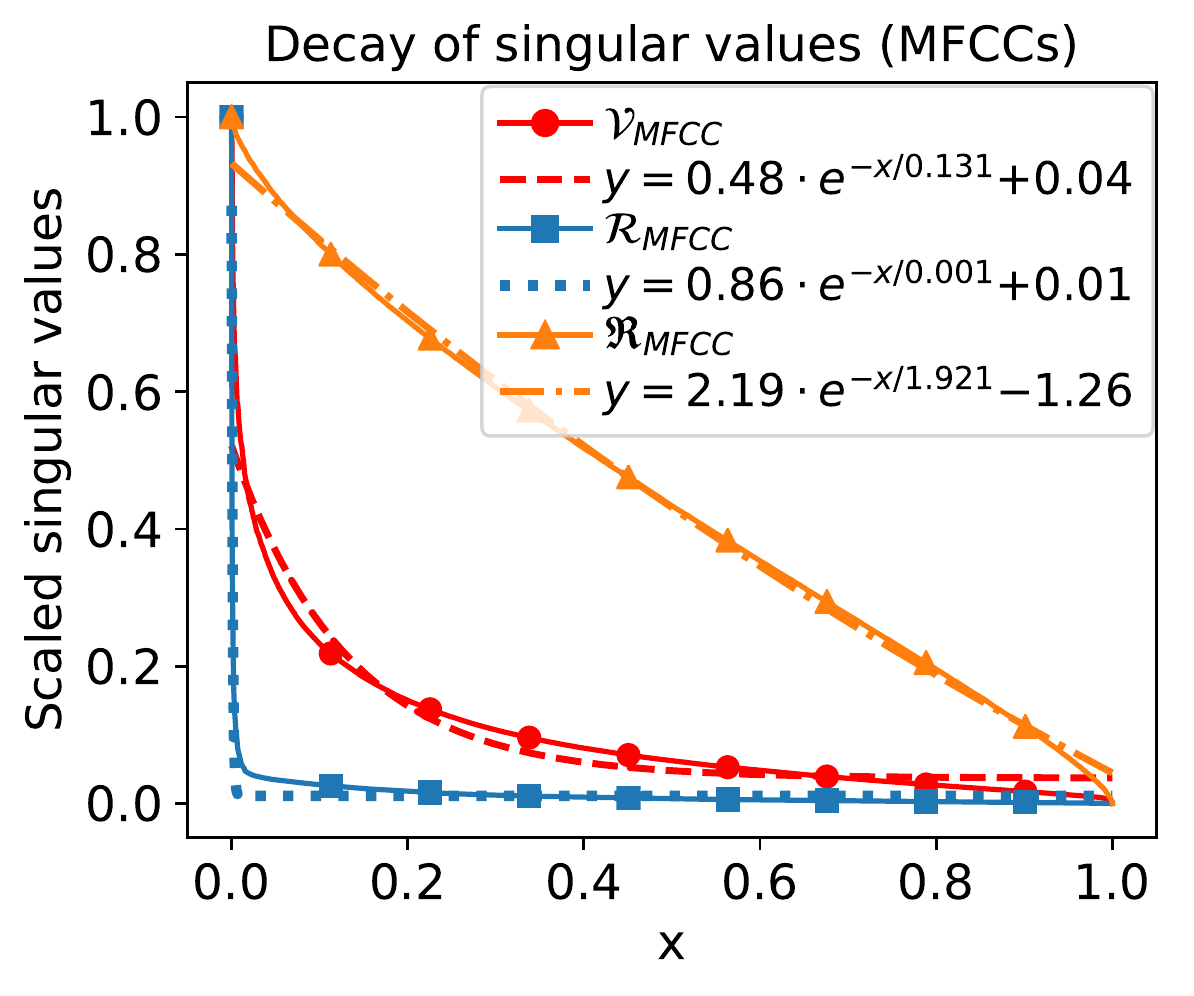}
    \caption{Left column: singular values obtained in the SVD of individual adversarial perturbations and random perturbations, computed in three feature representations: raw audio waveforms (top), spectrograms (center) and MFCCs (bottom). Right column: characterization of the decay of the singular values by fitting an exponential curve (the values in both axes have been scaled in the range [0,1]).}
    \label{fig:singular_values}
\end{figure}

We start evaluating the fooling rate of randomly sampled perturbations in the subspace spanned by the first $N=\{10,50,100,200,500\}$ singular vectors, for the cases in which the SVD is computed for individual perturbations ($\mathcal{V}_\text{MFCC}$) and random perturbations ($\mathcal{R}_\text{MFCC}$). All the sampled perturbations were normalized, and the fooling rate was evaluated for different scaling factors under the $\ell_2$ norm, in the range $[-200, 200]$. Note that, given an unit vector $v$, for any scalar $c\in\mathbb{R}$, $||v\cdot c||_2=|c|$. For reference, the median $\ell_2$ norm of the perturbations (in the MFCC) produced by the 10 universal attacks generated in Section \ref{sec:dominant_classes_in_our_domain}, measured in the test set, is approximately $100$.

\begin{figure}
    \centering
    \hspace{0.2cm}
    \includegraphics[scale=0.49]{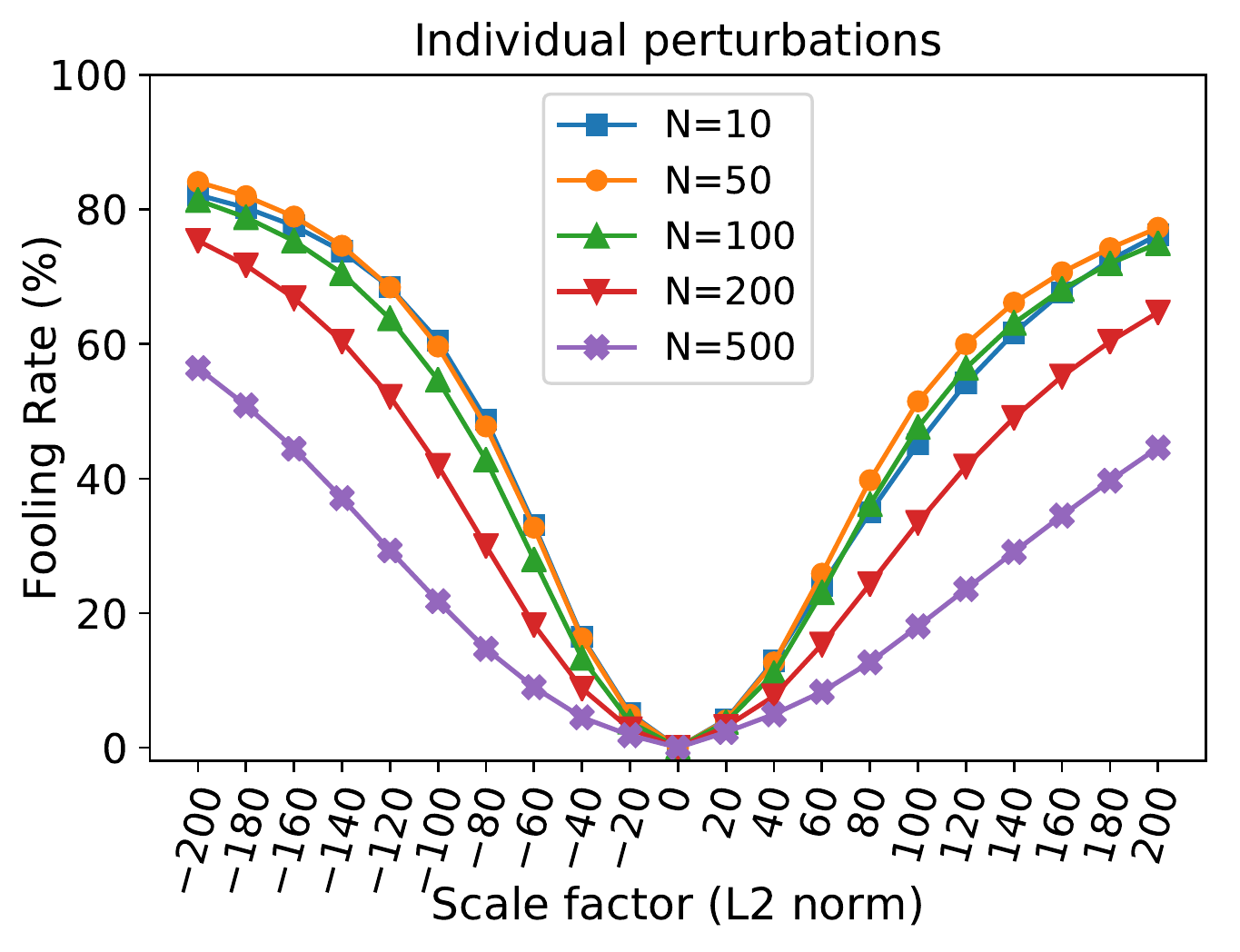}
    \includegraphics[scale=0.49]{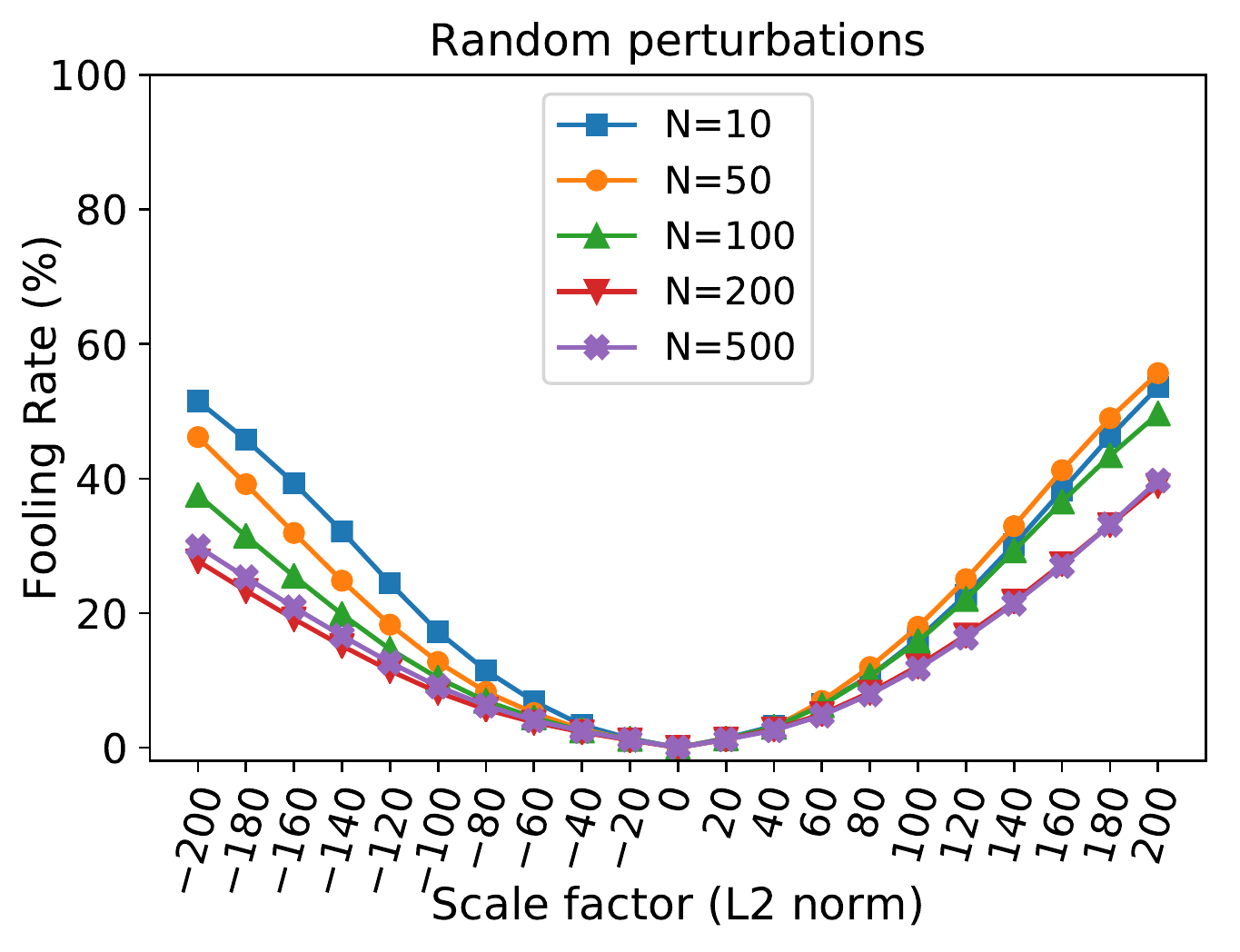} 
    \\
    \hspace{0.2cm}
    \includegraphics[scale=0.48]{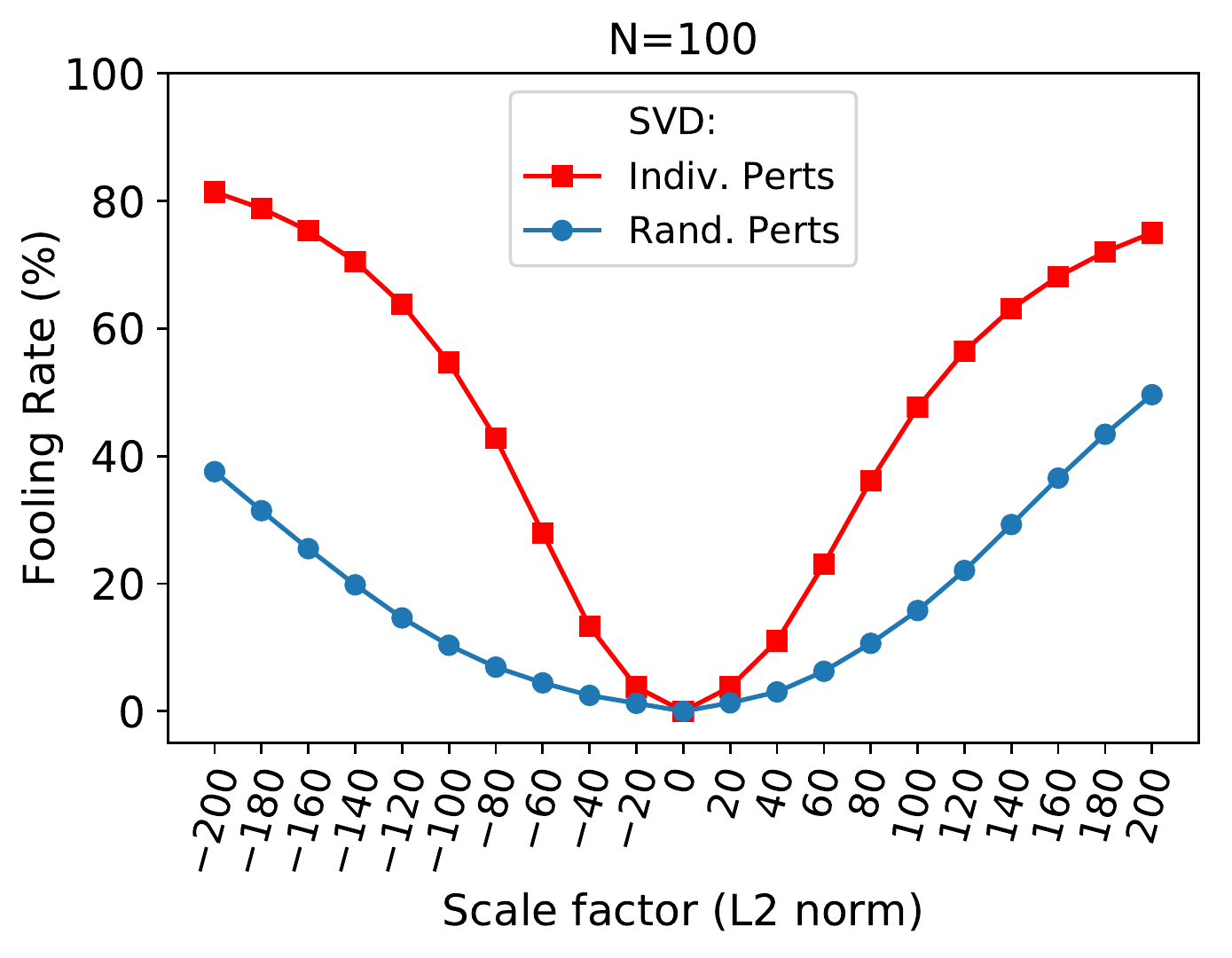} 
    \includegraphics[scale=0.48]{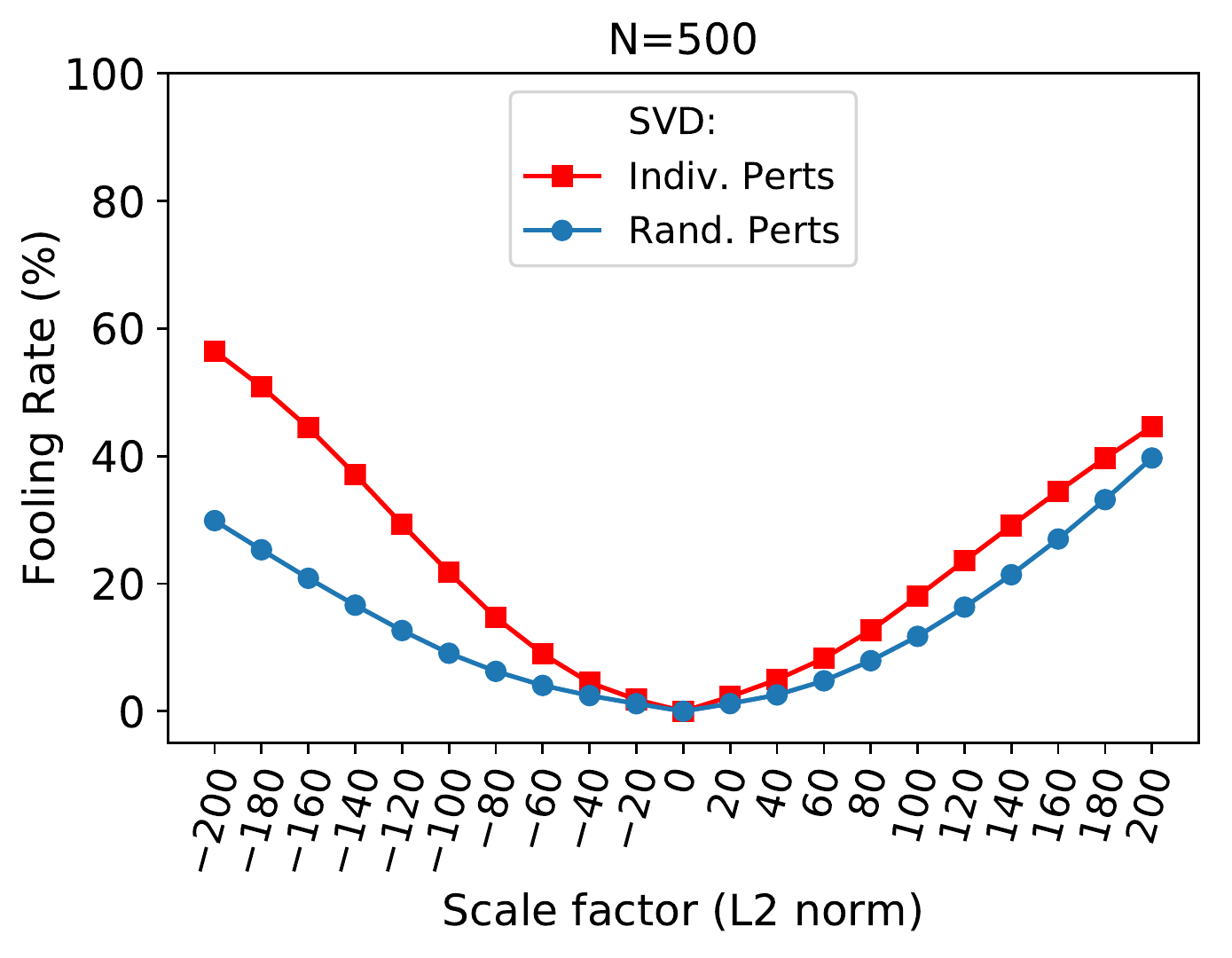}
    \caption{Fooling rate produced by random perturbations sampled from the subspace spanned by the first $N$ singular vectors. The results are averaged for 100 random perturbations. Each perturbation $v$ was normalized and multiplied by different scale factors $s_f$ (horizontal axis), so that $||v||_2=|s_f|$. The SVD is computed for individual perturbations (top left) and for random perturbations (top right), in the MFCC feature space. The bottom row shows a direct comparison between the average effectiveness of individual and random perturbations for $N=100$ (bottom left) and $N=500$ (bottom right).}
    \label{fig:svd_sampling_S}
\end{figure}

Figure \ref{fig:svd_sampling_S} shows the average fooling rates obtained for 100 trials,  for each value of $N$. The results clearly show that, when the SVD is computed for individual perturbations, the fooling rates are significantly higher than for the case of random perturbations, even for norms close to zero. For instance, taking as reference the results corresponding to an $\ell_2$ norm of 100, the average fooling rate is approximately 48\% for the case of individual perturbations, when $N\leq 100$. For the case of random perturbations, in the same conditions, the average fooling rate is only 17\%.

However, the fooling rate corresponding to individual perturbations significantly decreases when a large number of singular vectors are considered. Indeed, for $N\geq 200$, the fooling rates get closer to those obtained for random perturbations. For instance, when $N=500$, the average fooling rate (with an $\ell_2$ norm of 100) is approximately 18\%. This reveals that, whereas the singular vectors corresponding to the highest singular values are capturing directions normal to the decision boundaries around natural inputs (being, therefore, effective in fooling the model for a large number of inputs), the remaining singular vectors do not provide additional or relevant information.

\subsubsection{Connection with dominant classes}
In the previous section, we have shown that, also for speech command classification models, it is possible to find a low dimensional subspace $S$ containing (\textit{input-agnostic}) vectors normal to the decision boundaries in the vicinity of natural inputs.  Therefore, a reasonable hypothesis is that dominant classes can be explained in terms of the geometric similarity of the decision boundaries in regions surrounding natural inputs, information that is captured by the basis of $S$, that is, by the singular vectors obtained from the SVD of individual perturbations.

\begin{figure}
    \centering
    \includegraphics[scale=0.42]{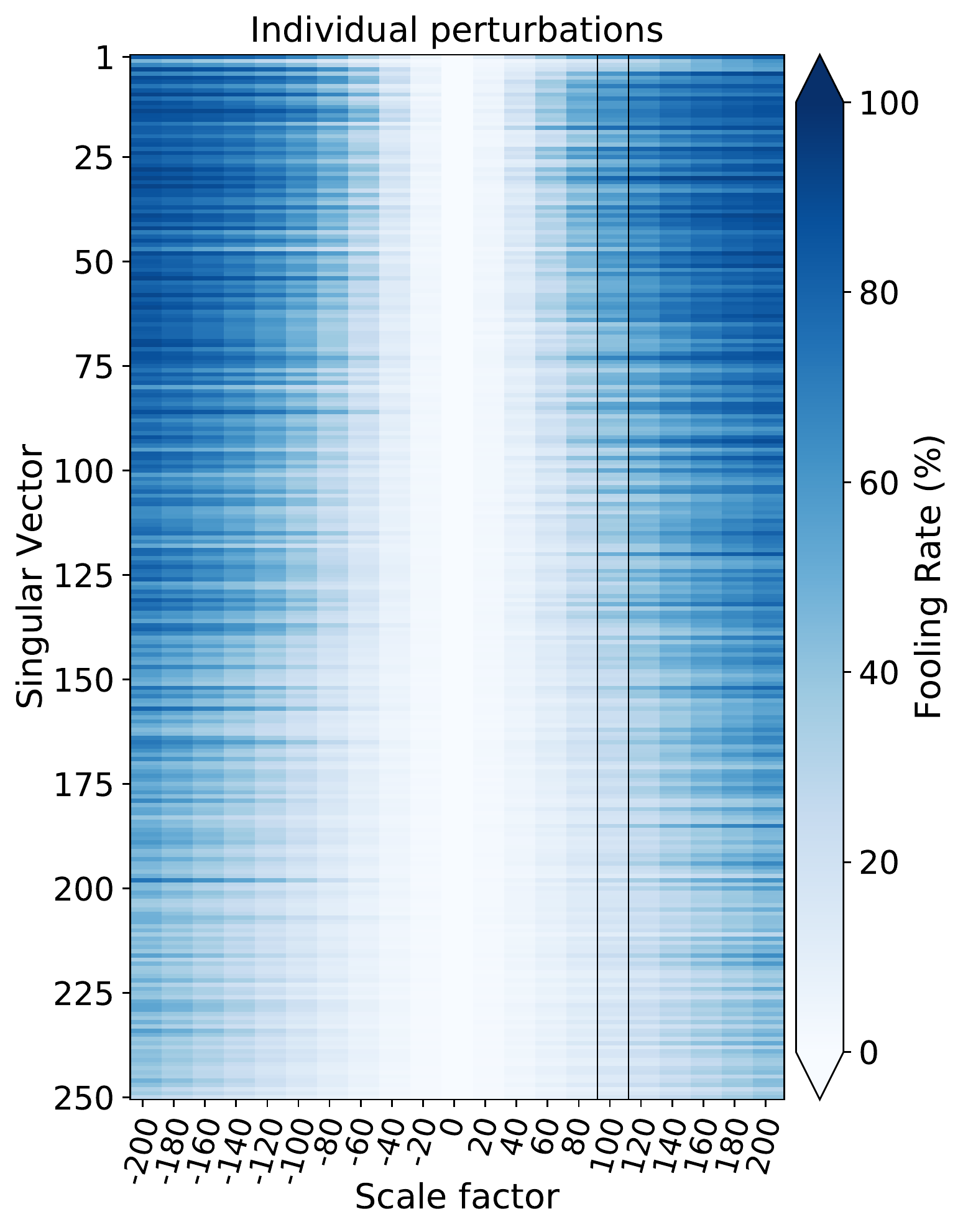} 
    \ \ \ 
    \includegraphics[scale=0.42]{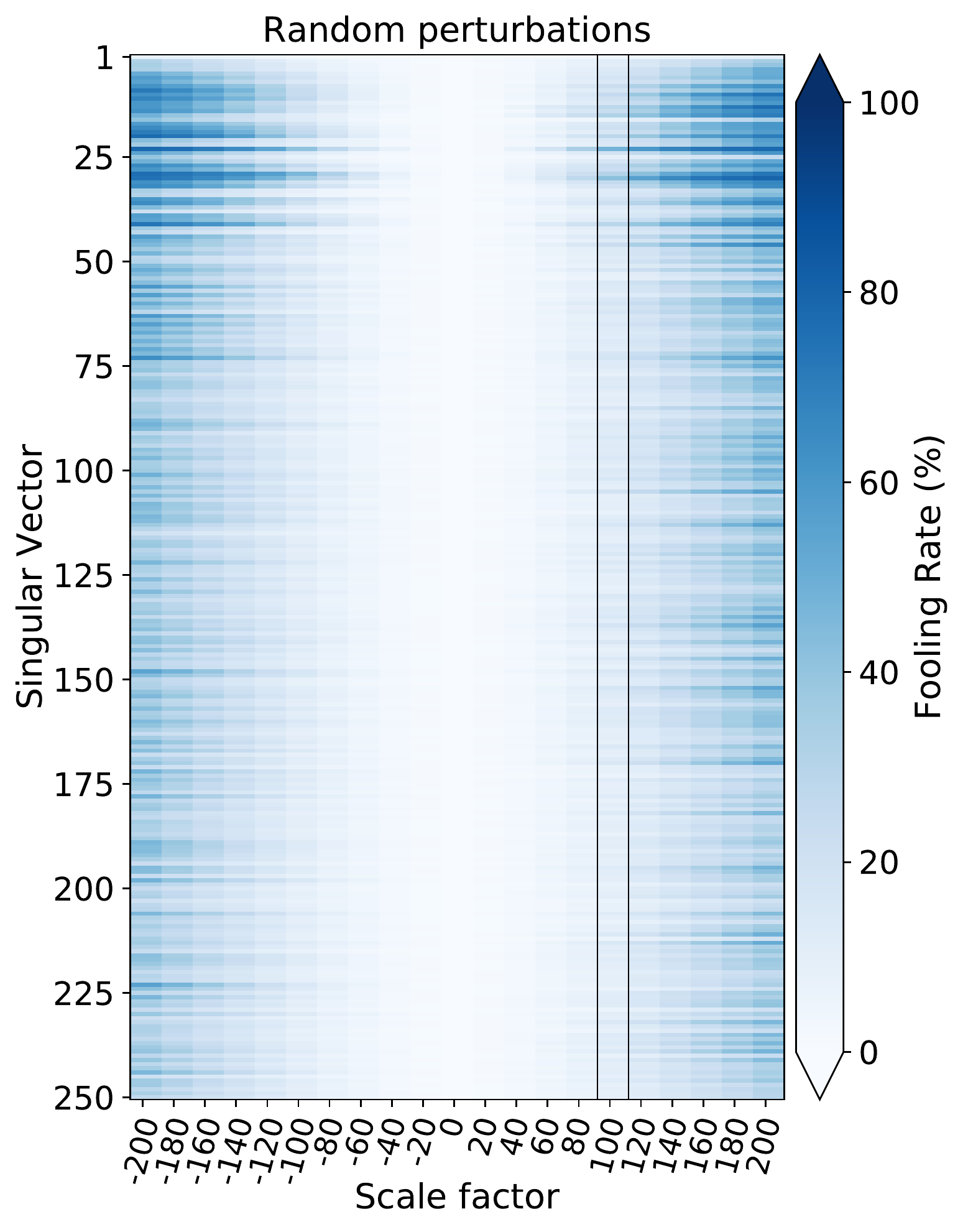}\\
    \hspace{-0.8cm}
    \includegraphics[scale=0.46]{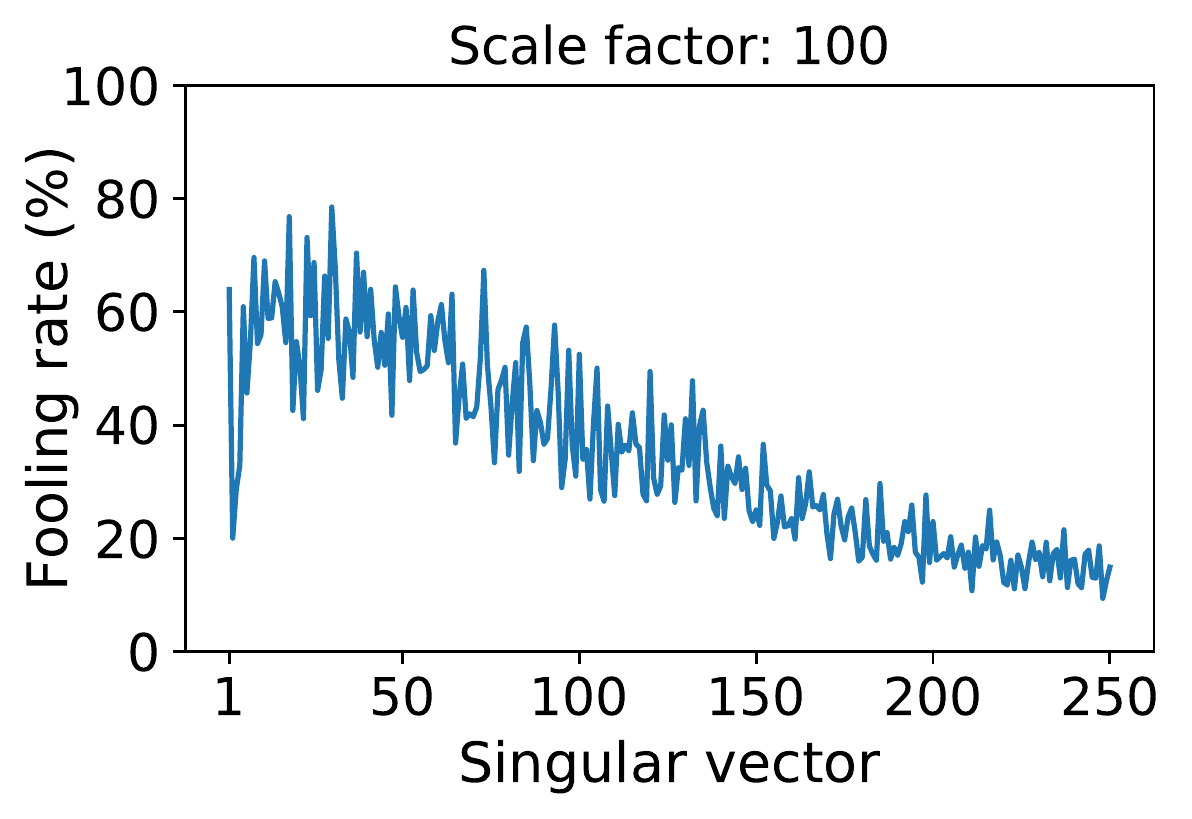} 
    \hspace{1.2cm}
    \includegraphics[scale=0.46]{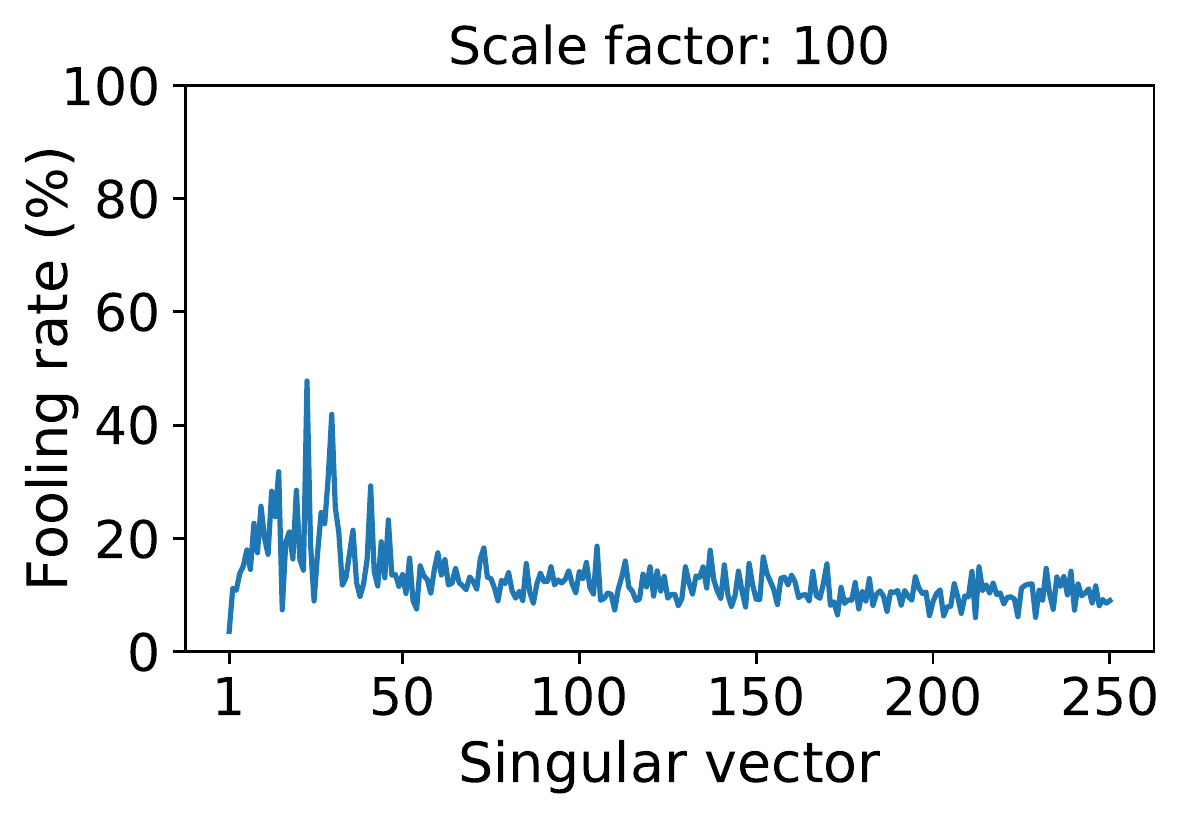} 
    \caption{Fooling rate percentage achieved when the inputs are perturbed with the first singular vectors computed for individual perturbations (left column) and for random perturbations (right column), in the MFCC feature space.}
    \label{fig:fr_sv_directions}
\end{figure}

The first hypothesis is that the first singular vectors are also normal to decision boundaries corresponding to the dominant classes. To validate this hypothesis, we first computed the fooling rate that each singular vector can achieve individually. This is shown in Figure \ref{fig:fr_sv_directions} (top left), in which the fooling rate of the first 250 singular vectors is reported for different $\ell_2$ norms. For reference, the results corresponding to a norm of 100 are also shown independently in the bottom-left part of the figure. The results clearly show that the first singular vectors are capable of fooling the model for a significant number of test inputs, particularly for the first 50 vectors, approximately. These fooling rates are also significantly higher than the ones obtained when the SVD is computed for random perturbations, which are also shown in Figure \ref{fig:fr_sv_directions} (right column).

To continue with the analysis, we computed the frequency with which each class is (wrongly) predicted, considering only the inputs that were misclassified when the singular vectors were used as perturbations. The aim of this analysis is to assess if there exists a direct connection with the dominant classes. The results are shown in Figure \ref{fig:adv_freqs_sv}, considering the first 100 singular vectors, scaled to have an Euclidean norm of $100$. As can be seen, considering the singular vectors with the highest fooling rate (those corresponding to the vectors approximately in the range [1,50]), the most frequent wrong classes are \textit{unknown} and \textit{left}. Indeed, for 84\% of the singular vectors in [1,50], the sum of the frequency corresponding to those two classes exceeds 50\%, that is, at least 50\% of the misclassified inputs are classified as \textit{left} or as \textit{unknown}. Moreover, for 62\% of the singular vectors, the total frequency corresponding to those two classes exceeds 80\%. Therefore, we now know that the singular vectors (with a high fooling rate) not only point towards decision boundaries in the close vicinity of natural inputs, but also that those decision boundaries correspond mainly to the dominant classes.

We repeated the experiment using the singular vectors obtained when the SVD is computed for random perturbations. The results are shown in Figure \ref{fig:adv_freqs_sv_rand}. In this case, it is evident that the results are more uniform along all the singular vectors, particularly for those singular vectors with a higher fooling rate (precisely, those in the range [1,50], as shown in Figure \ref{fig:fr_sv_directions}). For reference, in this case, only for 32\% of the singular vectors in the range [1,50] the total frequency corresponding to \textit{unknown} or \textit{left} exceeds 50\%, and only for 2\% of the singular vectors the total frequency exceeds 70\%.

Overall, the SVD of individual perturbations has shown that the obtained singular vectors are input-agnostic perturbations directions for which the model is highly vulnerable: even when the inputs are slightly pushed in those directions, they surpass the decision boundary of the model. This reveals that the geometry of the decision boundary has \textit{patterns} that are repeated in the vicinity of multiple natural inputs. Apart from that, we have shown that such \textit{adversarial} directions mainly point towards the decision boundaries corresponding to the dominant classes. Therefore, it can be concluded that the universal perturbation optimization algorithms implicitly exploit the \textit{shared} geometric patterns of decision boundaries to increase the effectiveness of the perturbations, leading to the same dominant classes in the majority of the cases.

\newcommand\tmpscale{0.62}
\begin{figure*}
    \centering
    \includegraphics[scale=\tmpscale]{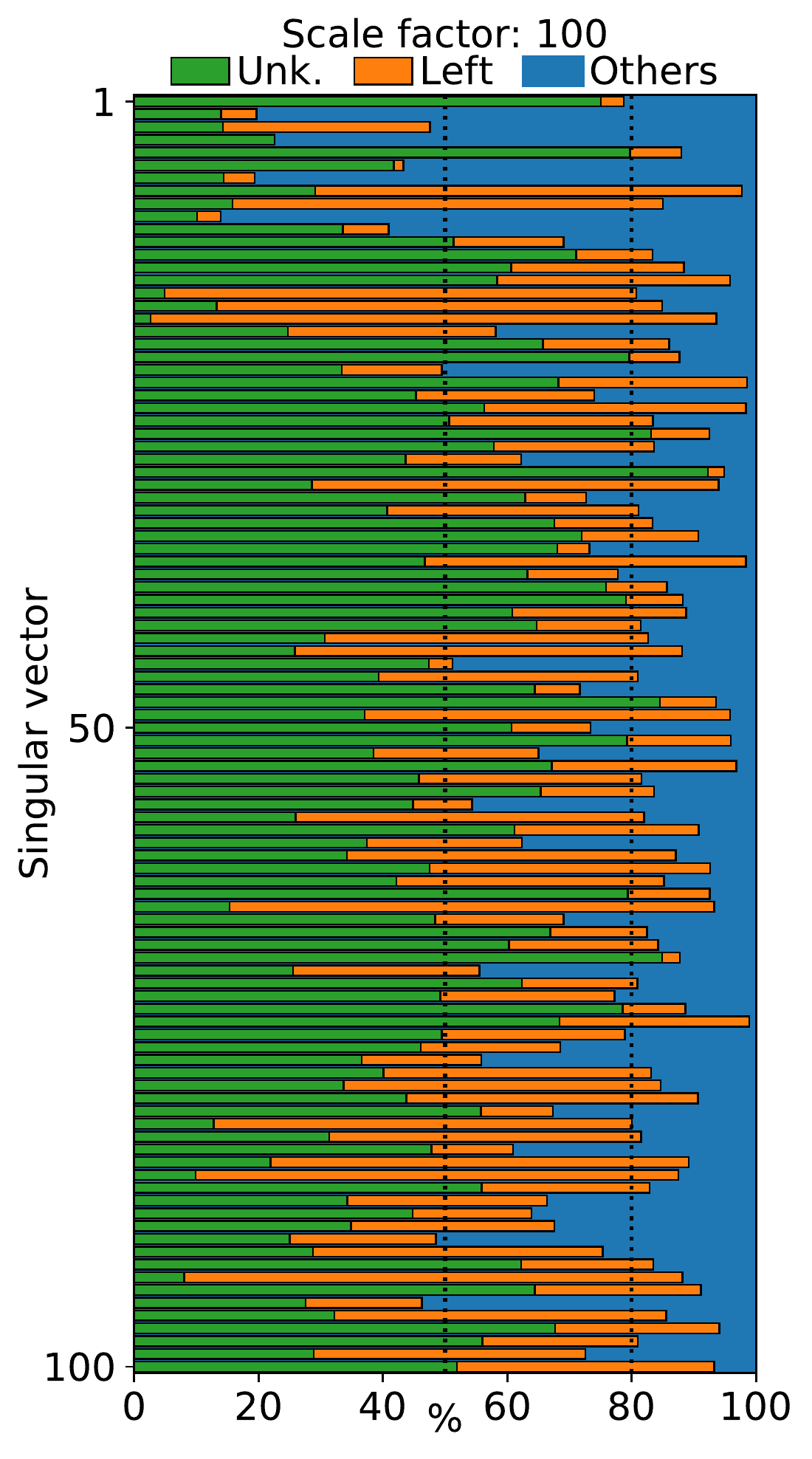}
    \includegraphics[scale=\tmpscale]{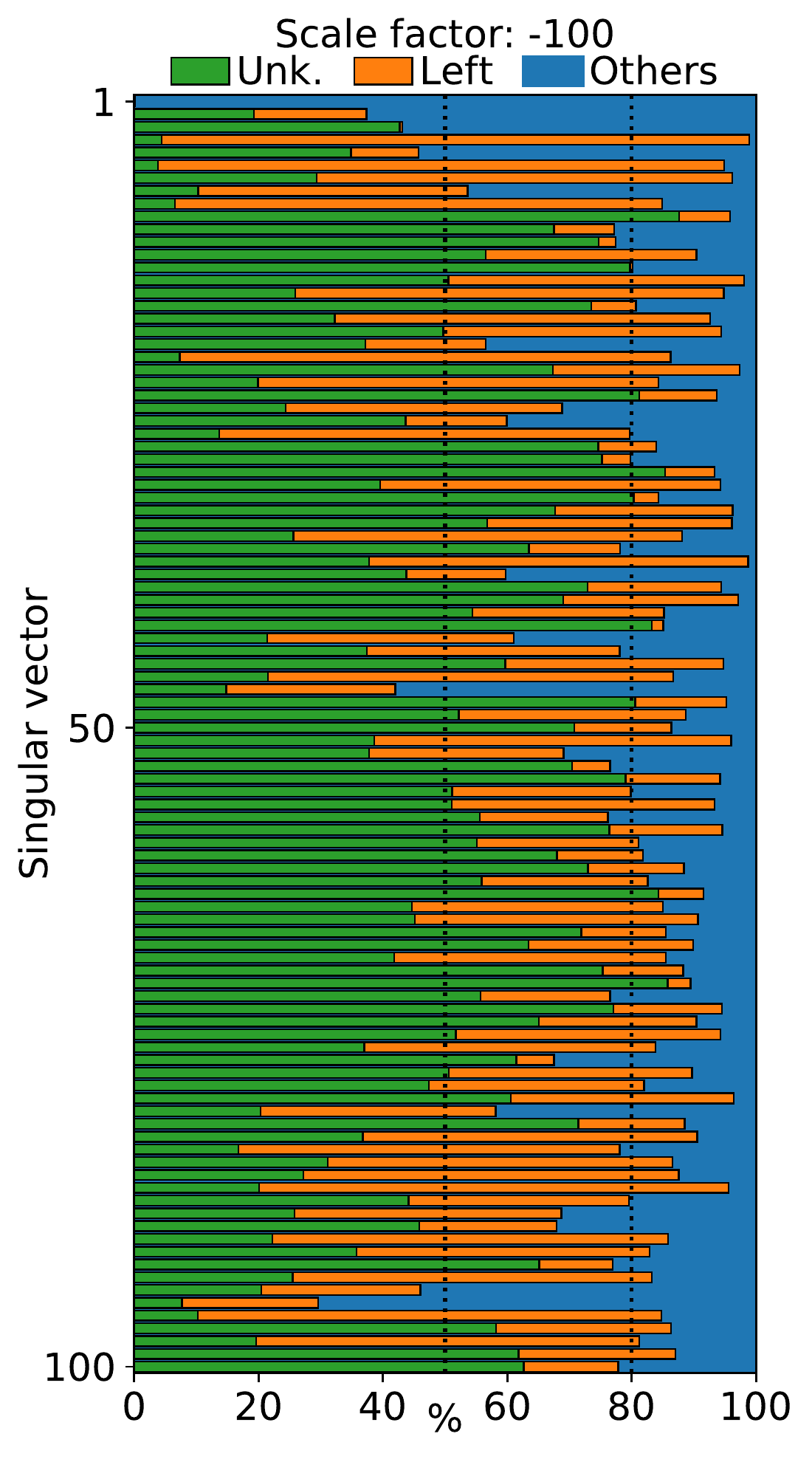}
    \caption{Frequency with which each class is assigned to the misclassified inputs under the effect of singular vectors (computed for \textbf{individual perturbations}, see Equation \eqref{eq:svd_indiv_mfcc}). The (unit) singular vectors have been scaled using two different scale factors: $100$ (left) and $-100$ (right). For the sake of clarity, the frequencies are shown individually for the classes \textit{unknown} and \textit{left}, while the total frequency corresponding to the rest of classes has been grouped (\textit{others}).}
    \label{fig:adv_freqs_sv}
\end{figure*}

\begin{figure*}
    \centering
    \includegraphics[scale=\tmpscale]{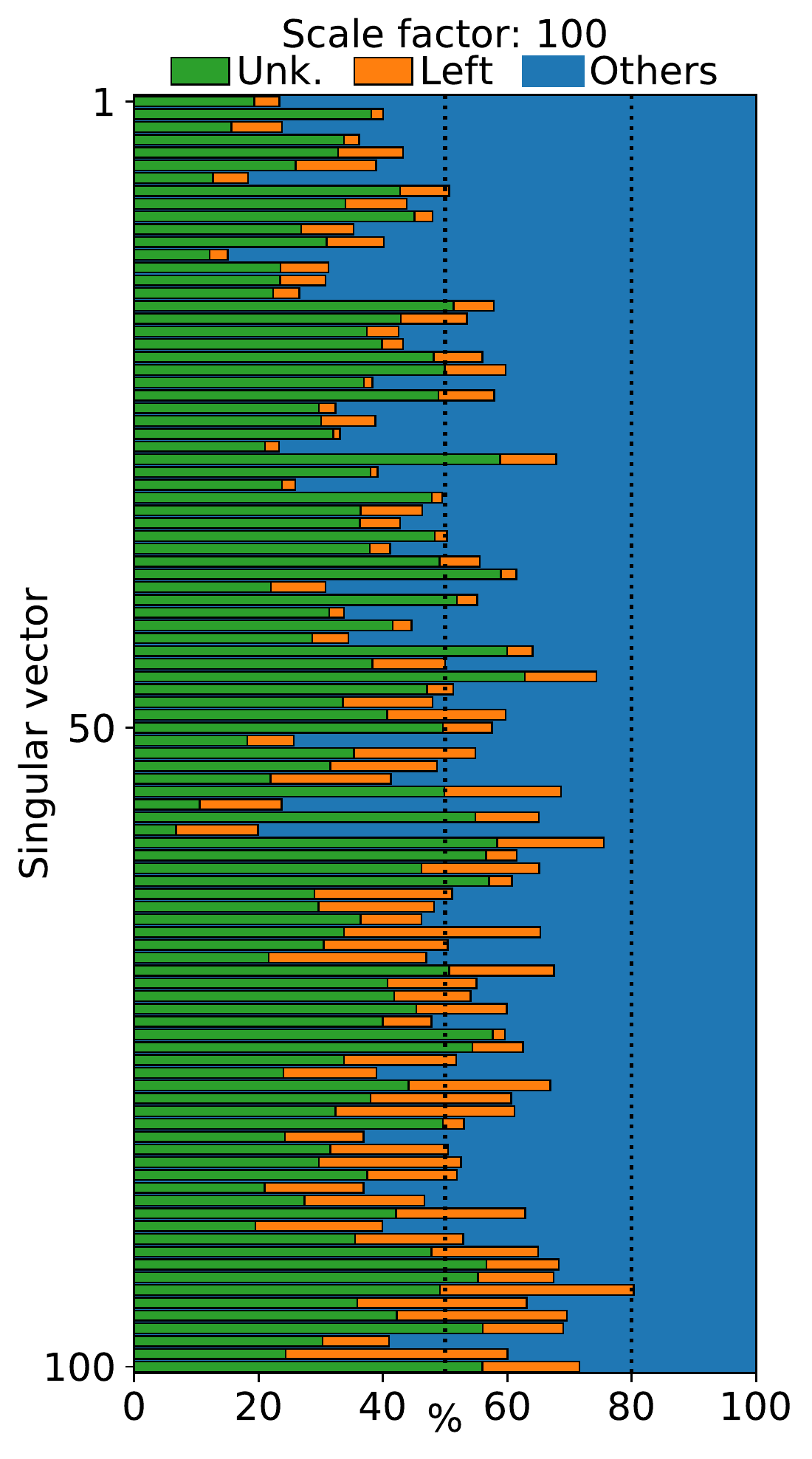}
    \includegraphics[scale=\tmpscale]{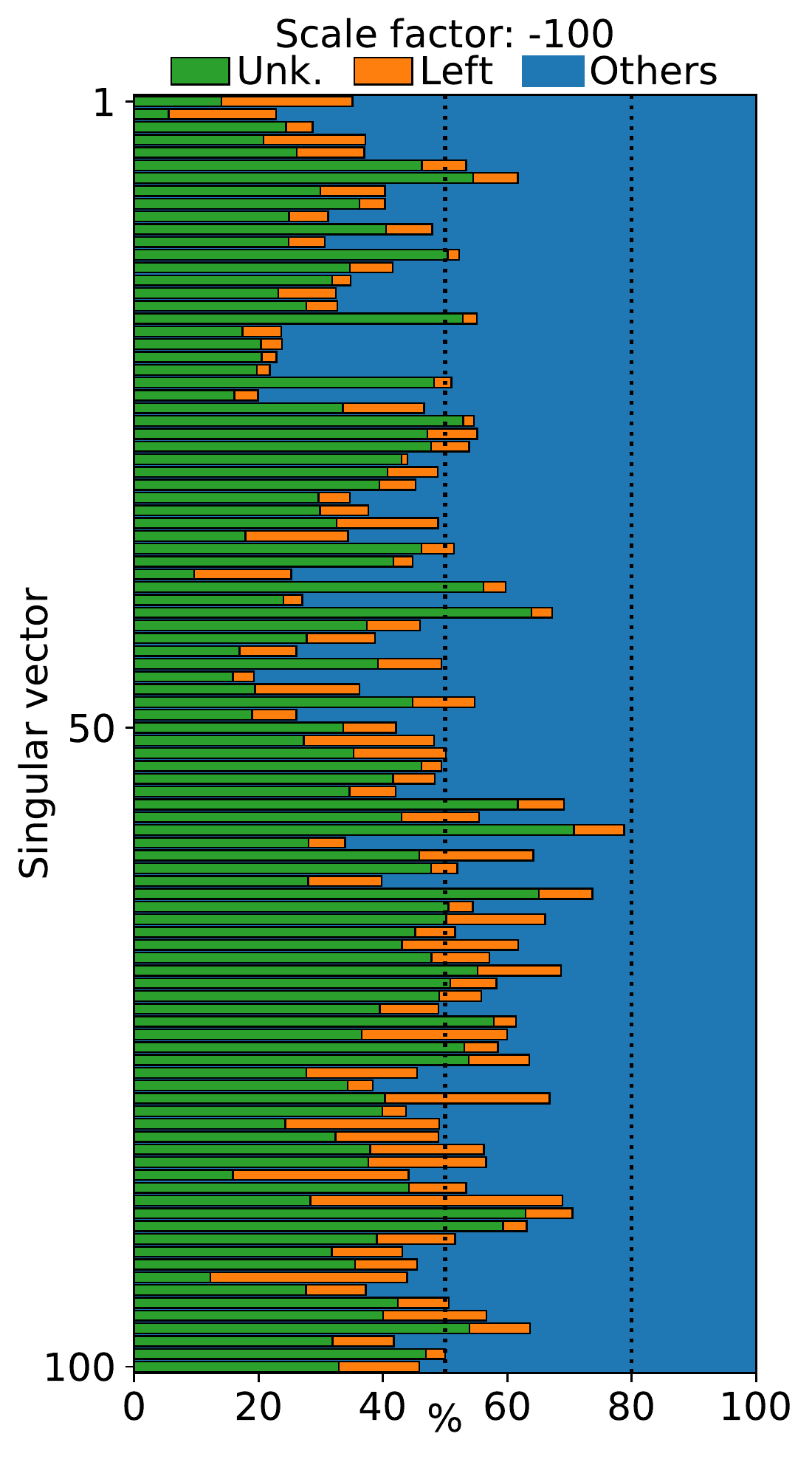}
    \caption{Frequency with which each class is assigned to the misclassified inputs under the effect of singular vectors (computed for \textbf{random perturbations}, see Equation \eqref{eq:rand_mfcc}). The (unit) singular vectors have been scaled using two different scale factors: $100$ (left) and $-100$ (right). For the sake of clarity, the frequencies are shown individually for the classes \textit{unknown} and \textit{left}, while the total frequency corresponding to the rest of classes has been grouped (\textit{others}).}
    \label{fig:adv_freqs_sv_rand}
\end{figure*}

\section{Conclusion}
\label{sec:conclusion}
In this paper, we have proposed and experimentally validated a number of hypotheses to justify the intriguing phenomenon of why universal adversarial perturbations for DNNs are capable of sending the majority of inputs towards the same wrong class (i.e., dominant classes), even if such behaviour is not specified during the optimization of the perturbations. These hypotheses were studied in the audio domain, using a speech command classification task as a testbed. To the best of our knowledge, previous work has examined this effect only in the image domain, proposing open explanations that we revisit. The results obtained from our analysis revealed multiple interesting facts regarding the vulnerability of DNNs to adversarial perturbations. On the one hand, we have shown that universal perturbations can be created just by optimizing a perturbation to be recognized by the model as one particular class with high confidence. This establishes a new perspective to create universal perturbations, while explains that a class is dominant if it contains patterns in the data distribution for which the model has a higher sensitivity. On the other hand, we demonstrated that the geometry of the decision boundaries of audio DNNs contains similar patterns in the vicinity of natural inputs, and that the most \textit{vulnerable} directions in the decision space point to the regions corresponding to the dominant classes. Finally, our work highlights a number of differences between the image domain and the audio domain, which contribute to a better and more general understanding of the field of adversarial machine learning.

\section{Future research lines}
\label{sec:future_lines}
Whereas the frameworks proposed in this paper have shown to be effective in revealing the connections between dominant classes and universal perturbations, there are a number of open lines that could be further investigated in order to achieve a deeper understanding of the behavior of universal perturbations. 

First, focusing on the framework proposed in Section \ref{sec:class_properties}, an interesting future line of research could be trying to identify the data-features that the model recognizes as each class with high confidence, for instance, following the methodologies proposed in recent related works \cite{ilyas2019adversarial}. Similarly, the analysis of the geometry of the decision space carried out in Section \ref{sec:svd} could be further extended by considering the curvature of the decision boundaries, which has proven to be highly informative for the analysis of universal perturbations \cite{moosavi-dezfooli2017analysis,jetley2018friends}.
Moreover, it could be interesting trying to unify the data-feature perspective used in Section \ref{sec:class_properties} and the one used in Section \ref{sec:svd}, relying on the geometry of the decision space of the DNN. Finally, a deeper understanding of the decision spaces of DNNs is necessary to comprehensively explain why decision boundaries contain large geometric correlations around natural inputs, as well as many other fundamental questions regarding the learning process of DNNs. 

Advances in all these research lines could bring a deeper understanding of the vulnerability of DNNs to adversarial attacks, which can be used, for instance, to create more effective attacks. Indeed, as shown in Section \ref{sec:dominant_classes_in_our_domain}, the existence of dominant classes reduces the effectiveness of universal perturbations, since the fooling rate in the inputs of those classes is practically zero. Therefore, preventing the appearance of dominant classes during the generation of the perturbation can lead to more effective attacks. At the same time, understanding the vulnerabilities of DNNs to adversarial attacks also contributes to the generation of more effective defensive strategies, and, ultimately, more robust models.

\section{Acknowledgments}
This work is supported by the Basque Government (BERC 2018-2021 and ELKARTEK programs, IT1244-19, and PRE\_2019\_1\_0128 predoctoral grant), by the Spanish Ministry of Economy and Competitiveness MINECO (projects TIN2016-78365-R and PID2019-104966GB-I00) and by the Spanish Ministry of Science, Innovation and Universities (FPU19/03231 predoctoral grant). Jose A. Lozano acknowledges support by the Spanish Ministry of Science, Innovation and Universities through BCAM Severo Ochoa accreditation (SEV-2017-0718).

\bibliographystyle{unsrt}
\bibliography{references}

\newpage

\begin{appendices}

\counterwithin{figure}{section}
\counterwithin{table}{section}

\section{Illustration of local approximations of decision boundaries}
\begin{figure}[!h]
    \centering
    \includegraphics[scale=0.22]{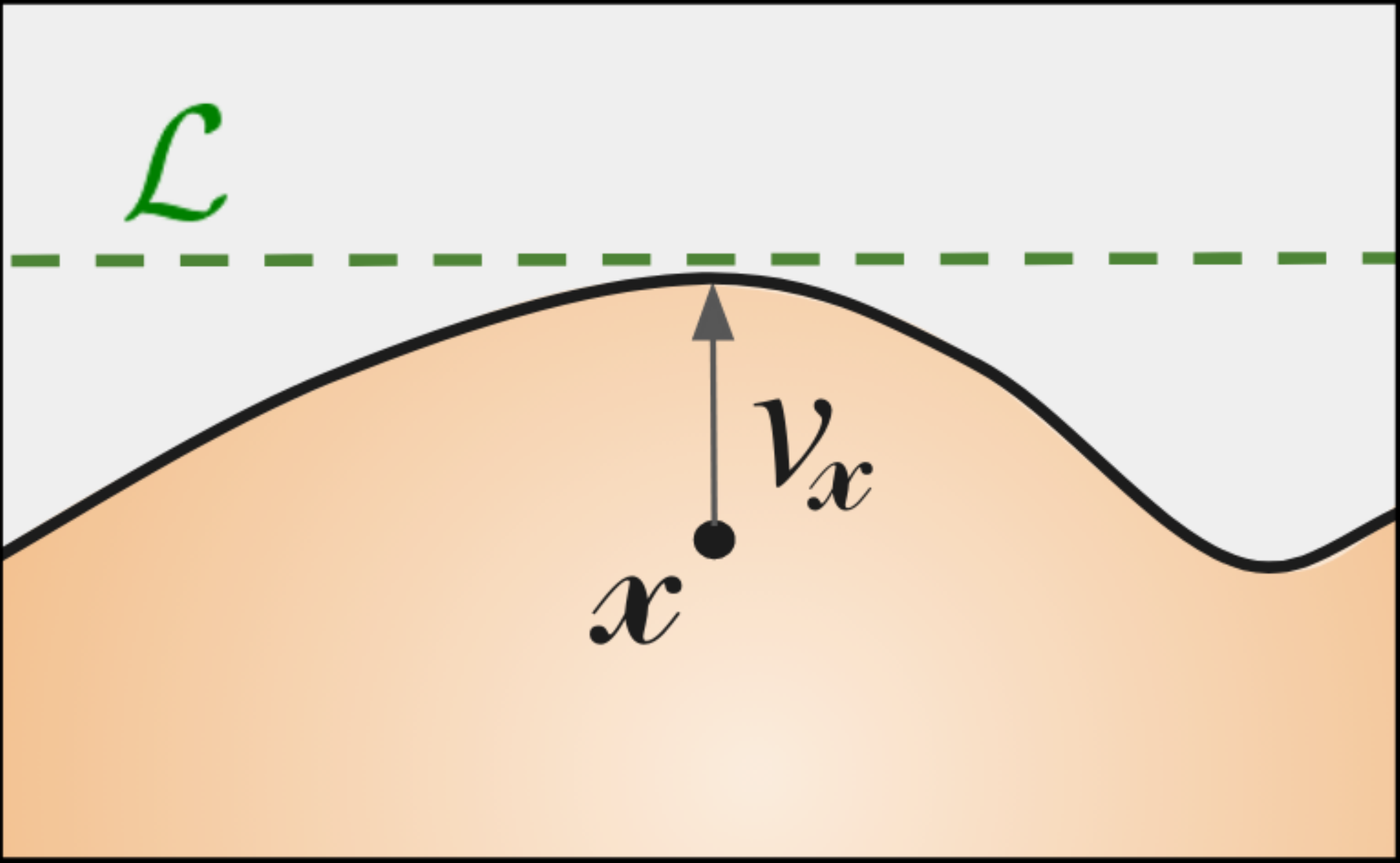} \hspace{0.2cm}
    \includegraphics[scale=0.22]{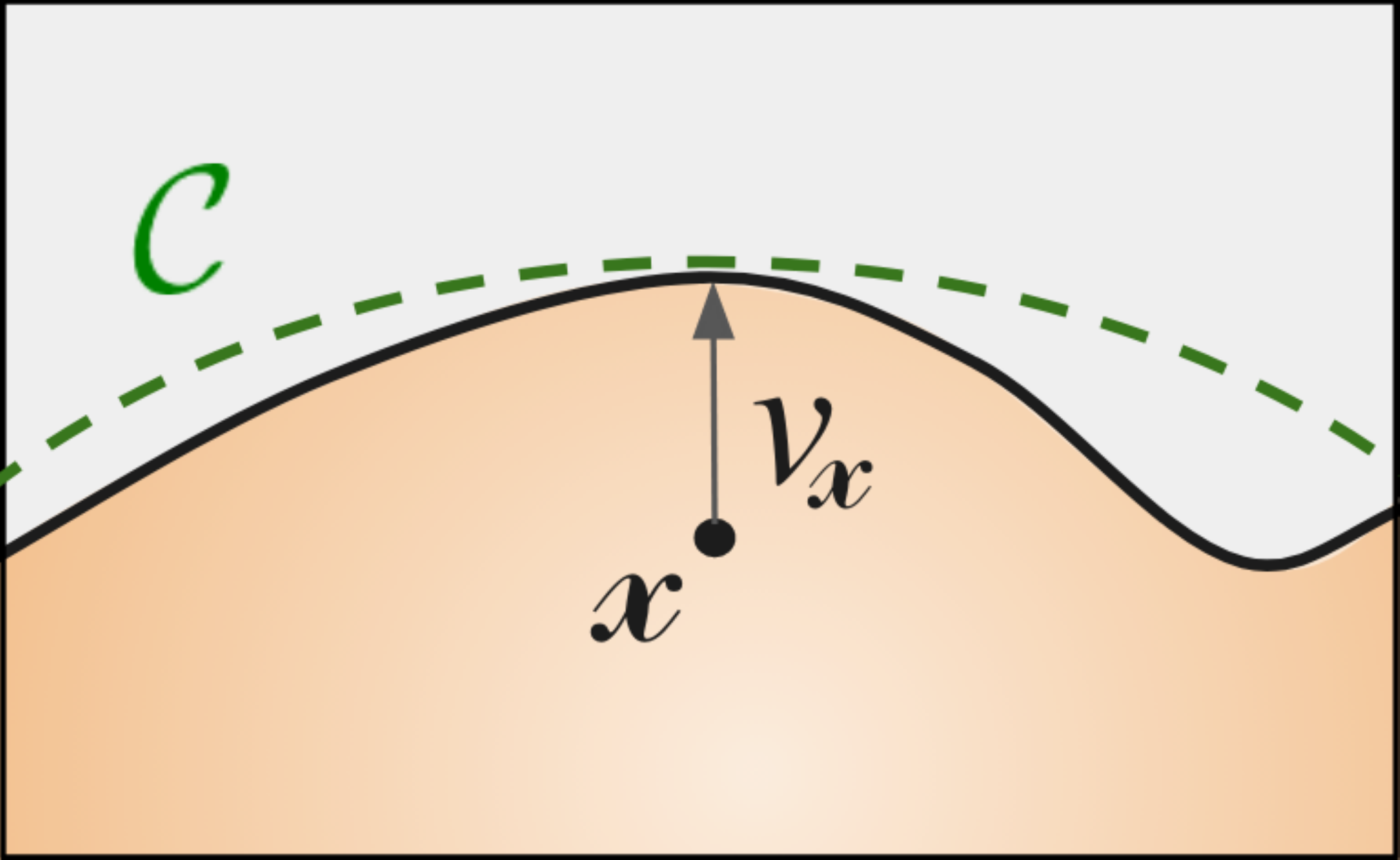} \hspace{0.2cm}
    \includegraphics[scale=0.189]{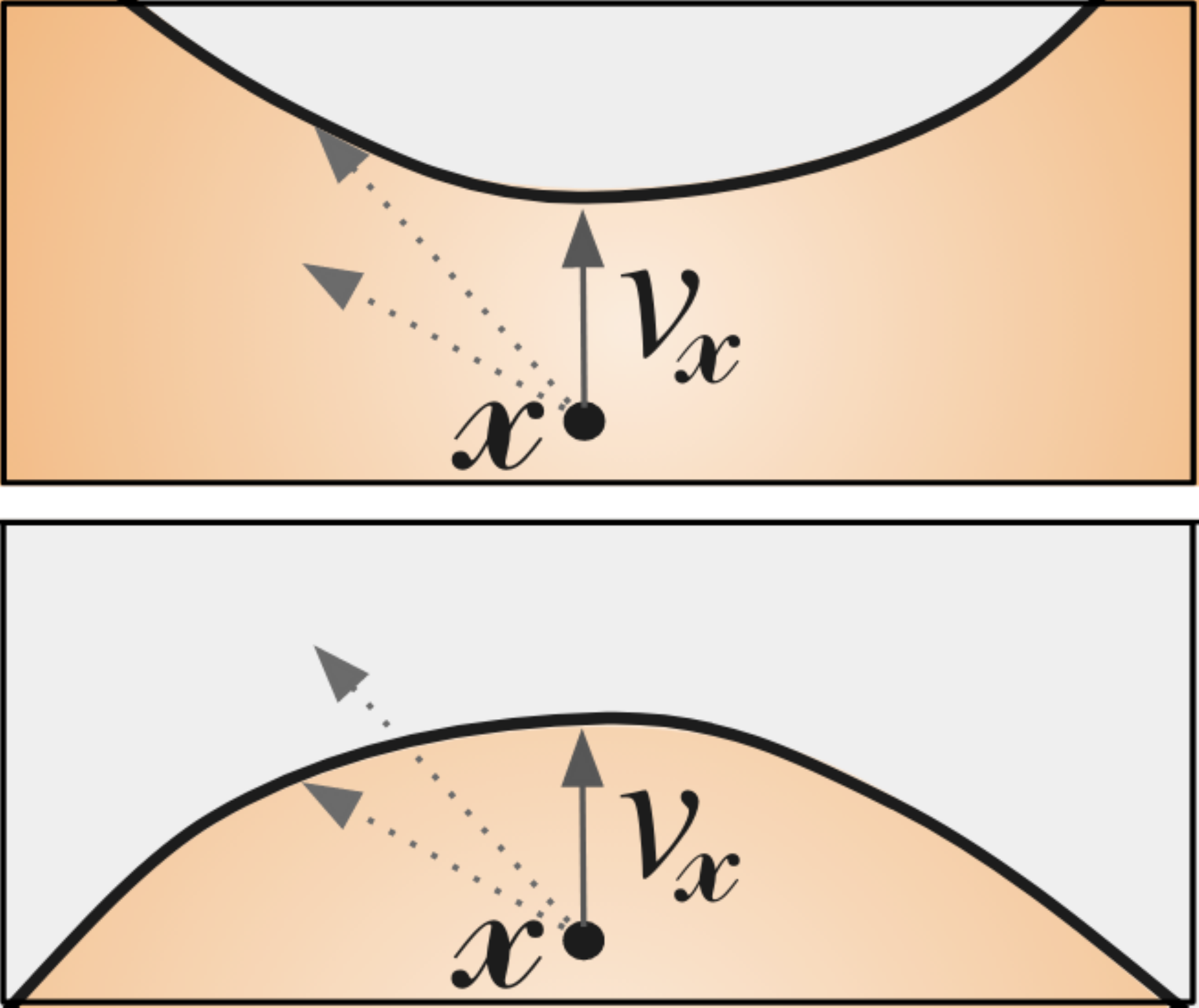}
    \caption{Illustration of the decision boundary approximations introduced in \cite{moosavi-dezfooli2017analysis}. The left image illustrates the locally linear (flat) decision boundary model, and the middle figure the locally curved decision boundary model. The solid curve corresponds to the actual boundary, and the dashed lines to the approximations. Note that in both cases the approximations are estimated at $x+v_x$, being $x$ an input sample and $v_x$ a vector \textit{normal} to the decision boundary (see Equation \ref{eq:optimal_adv}). The right images compare a positively curved boundary (bottom) with a negatively curved boundary (top) along $v_x$. Two dashed arrows have been included as reference in both images, to highlight that positively curved boundaries require smaller norms to be surpassed.}
    \label{fig:boundary_models}
\end{figure}

\section{Clean accuracy of the model in the test set}
\begin{table}[!h]
\centering
\begin{tabular}{@{}lcc@{}}
\toprule
Class & Accuracy & Samples \\ \midrule
\textit{Silence} & 99.51 & 408 \\
\textit{Unknown} & 66.42 & 408 \\
\textit{Yes} & 94.03 & 419 \\
\textit{No} & 74.57 & 405 \\
\textit{Up} & 92.00 & 425 \\
\textit{Down} & 80.79 & 406 \\
\textit{Left} & 89.81 & 412 \\
\textit{Right} & 88.64 & 396 \\
\textit{On} & 87.12 & 396 \\
\textit{Off} & 81.59 & 402 \\
\textit{Stop} & 93.67 & 411 \\
\textit{Go} & 77.36 & 402 \\ 
\midrule
Average & 85.52 & - \\
\bottomrule
\end{tabular}
\caption{Initial accuracy percentage of the DNN on the test set.}
\label{tab:test_clean_acc}
\end{table}

\newpage

\section{Detailed analysis of the effectiveness of universal perturbations (UAP-HC)}

Table \ref{tab:univ_fr_each_experiment} shows the effectiveness of each universal adversarial perturbation generated in Section \ref{sec:dominant_classes_in_our_domain}, using Algorithm \ref{alg:UAP-HC}.

\begin{table}[!h]
\centering
\begin{threeparttable}
\begin{tabular}{@{}@{\hskip 0.2in}lc@{\hskip 0.39in}c@{\hskip 0.3in}c@{\hskip 0.1in}}
\toprule
\multicolumn{1}{l}{\multirow{2}{*}{Experiment}} & \multicolumn{3}{c}{Restricted class} \\ \cmidrule(l){2-4} 
\multicolumn{1}{l}{}                            & None    & \{\textit{Left}\}  & \{\textit{Left,Unk.}\}  \\ 
\midrule
1 & 46.34 & 37.73 & 33.88 \\
2 & 35.29 & 31.56 & 34.24 \\
3 & 41.25 & 36.35 & 37.49 \\
4 & 38.47 & 37.42 & 34.91 \\
5 & 38.35 & 32.86 & 34.31 \\
6 & 30.13 & 30.30 & 29.84 \\
7 & 32.52 & 34.55 & 32.88 \\
8 & 33.98 & 34.29 & 30.94 \\
9 & 41.08 & 37.14 & 33.86 \\
10 & 41.94 & 36.80 & 35.15 \\
\midrule
Mean & 37.94 & 34.90 & 33.75 \\
Mean\tnote{1} & 41.68 & 37.39 & 37.08 \\
Mean\tnote{2} & 44.97 & 40.32 & 39.90 \\ \bottomrule
\end{tabular}
\begin{tablenotes}
\item[1] Without considering dominant classes.
\item[2] Without considering dominant classes and \textit{Silence}.
\end{tablenotes}
\end{threeparttable}
\caption{Fooling rate percentage of the universal adversarial perturbations generated using Algorithm \ref{alg:UAP-HC}. The results are computed for a set of \textit{test} samples, which were not seen during the generation of the universal perturbations.}
\label{tab:univ_fr_each_experiment}
\end{table}

\end{appendices}

\end{document}